\documentclass[final]{cvpr}
\usepackage{nopageno}
\makeatletter
\@namedef{ver@everyshi.sty}{} 
\makeatother
\usepackage[normalem]{ulem}
\usepackage{tikz}
\usetikzlibrary{patterns}

\usepackage{times}
\usepackage{tabularx}
\usepackage{epsfig}
\usepackage{graphicx}
\usepackage{amsmath}
\usepackage{enumitem}
\usepackage{booktabs}
\usepackage{gensymb}
\usepackage{amssymb}
\usepackage{multirow}
\usepackage{caption}
\usepackage{subcaption}
\usepackage{makecell}
 \usepackage[export]{adjustbox}
\usepackage[pagebackref=true,breaklinks=true,colorlinks,bookmarks=false]{hyperref}

\def\cvprPaperID{7086} %
\def\confYear{CVPR 2021}

\newcommand{\corr}[2]{#2}   %
\newcommand\tikzmark[2]{%
\tikz[remember picture,baseline] \node[above, outer sep=0pt, inner sep=0pt] (#1){\phantom{#2}};%
}

\newcommand\link[2]{%
\begin{tikzpicture}[remember picture, overlay, >=stealth, shift={(0,0)}]
  \draw[->] (#1) to (#2);
\end{tikzpicture}%
}
\begin{document}

\title{CoMoGAN: continuous model-guided image-to-image translation}

\author{Fabio Pizzati\\
Inria, Vislab\\
{\tt\small fabio.pizzati@inria.fr}
\and
Pietro Cerri\\
Vislab\\
{\tt\small pcerri@ambarella.com}
\and
Raoul de Charette\\
Inria\\
{\tt\small raoul.de-charette@inria.fr}
}

\maketitle

\begin{abstract}

CoMoGAN is a continuous GAN relying on the unsupervised reorganization of the target data on a functional manifold. To that matter, we introduce a new Functional Instance Normalization layer and residual mechanism, which together disentangle image content from position on target manifold. We rely on naive physics-inspired models to guide the training while allowing private model/translations features.
CoMoGAN can be used with any GAN backbone and allows new types of image translation, such as cyclic image translation like timelapse generation, or detached linear translation. On all datasets\corr{and metrics}{}, it outperforms the literature. Our code is available in this page:\\ \url{https://github.com/cv-rits/CoMoGAN}.
\end{abstract}

\vspace{-0.8em}
\section{Introduction}
Image-to-image (i2i) translation networks learn translations between domains, applying to the context of source images a target appearance learned from a dataset. This enables applications such as neural photo editing~\cite{zhu2017unpaired,liu2017unsupervised,huang2018multimodal,pumarola2020ganimation,chen2019homomorphic}, along with robotics-oriented tasks as time-of-day or weather selection~\cite{zheng2020forkgan,pizzati2019domain,pizzati2020model,gong2020analogical,tremblay2020rain}, domain adaptation~\cite{hoffman2017cycada,murez2018image,li2019bidirectional,toldo2020unsupervised}, or others. Despite impressive leaps forward with unpaired~\cite{zhu2017unpaired,liu2017unsupervised}, multi-target~\cite{choi2020stargan,wu2019relgan}, or continuous~\cite{wang2019deep,gong2019dlow} i2i, there are still important limitations. Specifically, to learn complex continuous translations existing works require supervision on intermediate domain points. Also, they assume piece-wise or entire linearity of the domain manifold. Such constraints can hardly meet cyclic translations (e.g. daytime) or continuous ones costly or impractical to label (e.g. fog, rain). 

\begin{figure}
	\centering
	\includegraphics[width=\linewidth]{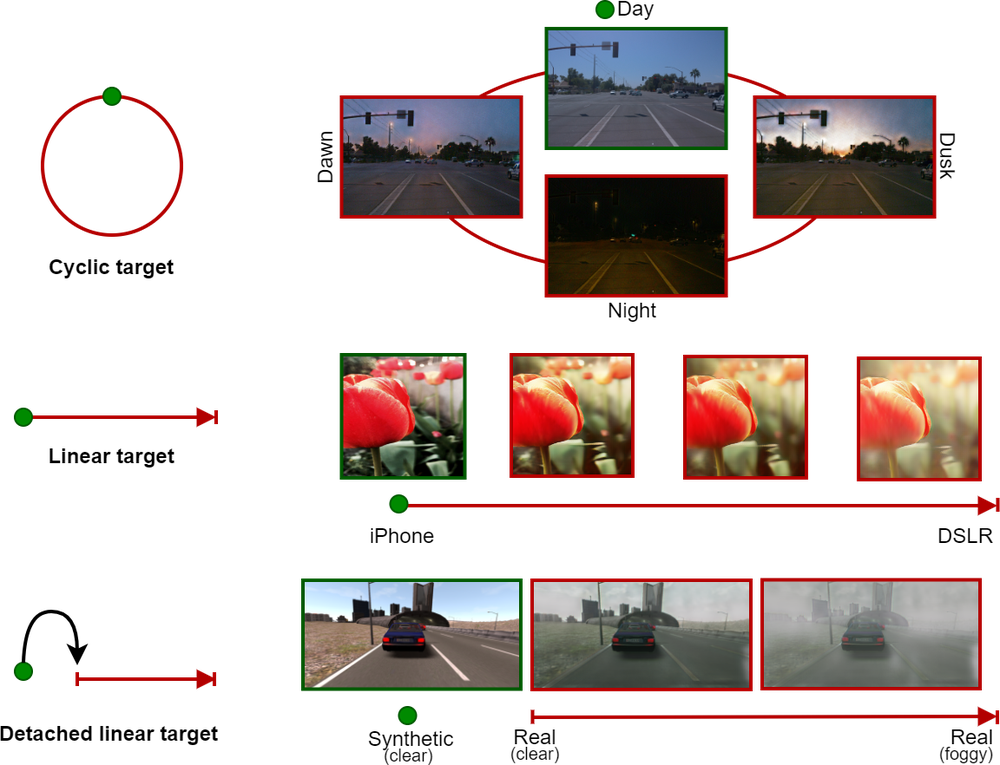}
	\caption{Detaching from traditional i2i translation, we are interested in \textit{continuous} mapping from source domain (green point) to a target domain (red lines), in single- or multi- modal setup.
		A key feature of our proposal, is unsupervised reorganization of the data along a functional manifold (top: cyclic, middle/bottom: linear). 
		We leverage lighting translations from day images (top), shallower depth of field from in-focus images (middle), or synthetic clear images to realistic foggy images (bottom).}\vspace{-0.5em}
	\label{fig:teaser}
\end{figure}

Instead, we introduce CoMoGAN, the first i2i framework learning non-linear continuous translations with unsupervised target data. It is trained using simple physics-inspired models for guidance, while relaxing model dependency via continuous disentanglement of domain features. An interesting resulting property is that CoMoGAN discovers the target data manifold ordering, unsupervised. 
For evaluation we propose new translation tasks, shown in Fig.~\ref{fig:teaser}, being either cyclic/linear, attached/detached from source. Our contributions are:
\begin{itemize}[noitemsep,topsep=0pt]
    \item a novel model-guided setting for continuous i2i,
    \item CoMoGAN: an unsupervised framework for disentanglement of continuously evolving features in generated images, using simple model guidance,
    \item a novel Functional Instance Normalization (FIN) layer,
    \item the evaluation of CoMoGAN against recent baselines and new tasks, outperforming the literature on all.
\end{itemize}

\begin{figure*}
	\centering
	\includegraphics[width=\linewidth]{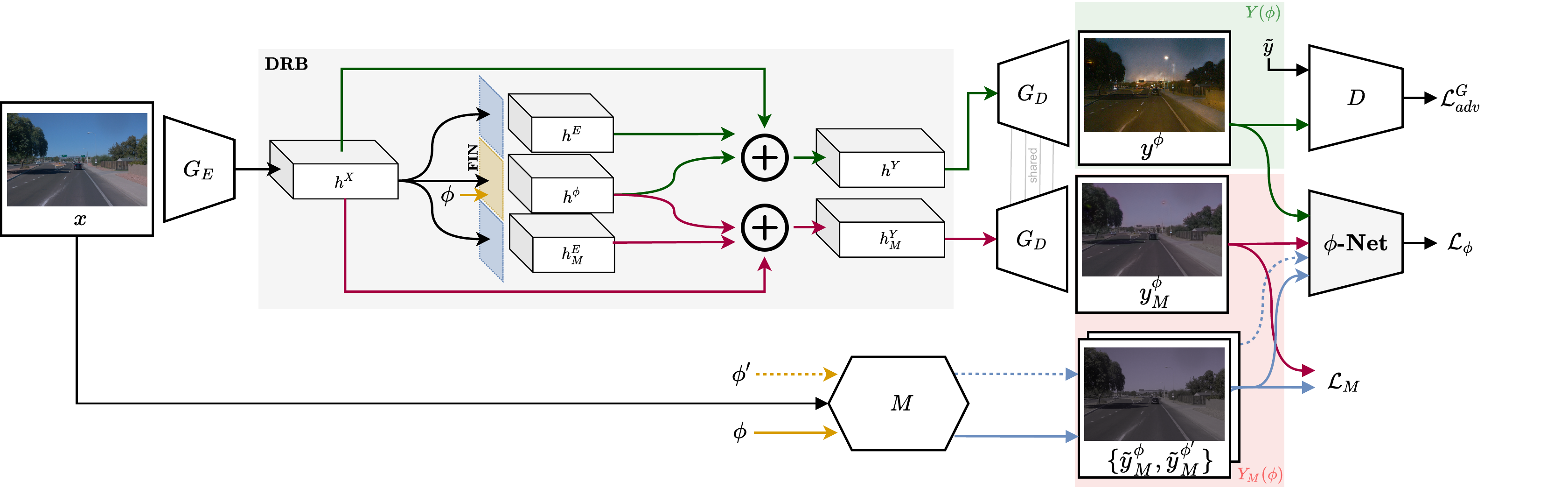}
	\caption{CoMoGAN enables unsupervised continuous translation, being end-to-end trainable, and architecture agnostic.
		Our Disentanglement Residual Block (DRB) -- placed between encoder/decoder ($G_E/G_D$) -- uses new Functional Instance Normalization (FIN, yellow layer) to learn manifold reshaping and continuous translation, guided with simple physics-inspired model $M$. 
		For losses~($\mathcal{L}$), on top of standard ones we optimize model reconstruction~($\mathcal{L}_M$) and manifold consistency~($\mathcal{L}_{\phi}$) by enforcing manifold distances between GAN output and model outputs $\{\phi, \phi'\}$ with a pair-wise estimator ($\phi\text{-Net}$).}
	\label{fig:method}\vspace{-0.06em}
\end{figure*}
\section{Related works}\label{sec:related}
Differently from early i2i~\cite{isola2017image}, the seminal work in \cite{zhu2017unpaired,yi2017dualgan} enabled unpaired source/target training. Building on it, multi-modal ~\cite{huang2018multimodal,zhu2017toward} or multi-target~\cite{choi2018stargan, choi2020stargan,wu2019relgan,almahairi2018augmented} i2i appeared. Performance was also boosted with additional supervision~\cite{shen2019towards,bhattacharjee2020dunit,mo2018instagan,li2018semantic,tang2020multi,cherian2019sem,zhu2020semantically,zhu2020sean,lin2020multimodal,ma2018exemplar,park2019semantic,musto2020semantically,lutjens2020physics}.

\paragraph{Model-guided translation.}
Models can be exploited to improve i2i. In~\cite{tremblay2020rain}, they hybrid a physics based rendering~\cite{halder2019physics} with GANs to enable controllable rainy translation. 
Similarly, \cite{pizzati2020model} disentangles occlusions by injecting models at training. 
All these rely on model integration, rather than guidance. 
Models could influence many training aspects, in the form of output space conditions~\cite{reichstein2019deep}, loss functions~\cite{Karpatne2017PhysicsguidedNN} or ad-hoc data augmentation~\cite{xie2018tempogan}. 
They have been used extensively for image restoration~\cite{pan2018physics,li2019heavy,yang2018towards}, but rarely for GAN image synthesis. Still, \cite{jahanian2019steerability} uses simple models to learn basic image transformation (rotation, brightness, etc.).

\paragraph{Disentangled representations.}
Disentanglement is commonly used to gain control on generation by separating image content and style ~\cite{huang2018multimodal,lee2019drit++,jiang2020tsit,park2020contrastive}. Others aim at controlling output images granularity~\cite{singh2019finegan} or specific features, as blur~\cite{lu2019unsupervised} or view-points~\cite{nguyen2019hologan}. Some exploit disentanglement for few-shot generalization capabilities~\cite{liu2019few,saito2020coco}. Domain features disentanglement also unifies representations across domains~\cite{xia2020unsupervised,lin2019exploring}. While some do not use labels at all~\cite{anokhin2020high,baek2020tunit}, none of them learn translation sequentiality.

\paragraph{Continuous image translation.}
A common practice for continuous i2i is to use intermediate domains by weighting discriminator~\cite{gong2019dlow,gong2020analogical}, using losses for middle states~\cite{wu2019relgan}, or mixing disentangled styles representations~\cite{choi2020stargan,romero2019smit}. 
Attribute vectors interpolation~\cite{xiao2017dna,zhang2019multi,mao2020continuous} enables continuous control of several features. Others continuously navigate latent spaces with discovered paths~\cite{chen2019homomorphic,goetschalckx2019ganalyze,jahanian2019steerability}. 
Finally, feature~\cite{upchurch2017deep} or kernel~\cite{wang2019deep} interpolation were proposed. 
Still, they assume linear interpolation -- not always valid (e.g. day to night include dusk). GANimation~\cite{pumarola2020ganimation} instead, use non-linear interpolations but require intermediate domain labels.

\section{CoMoGAN}
Instead of a point-to-point mapping ($X\mapsto{}Y$), CoMoGAN learns a continuous domain translation controlled by~$\phi$, that is $X\mapsto{}Y(\phi)$. Training uses source data (at fixed~$\phi_0$) and unsupervised target data (unknown~$\phi$). It reshapes the data manifold guided by naive physics-inspired models (e.g. tone-mapping, blurring, etc.). 
Rather than mimicry, we relax the model and let the networks discover private image features via our disentanglement of output, $\phi$, and style. 

Fig.~\ref{fig:method} is an overview of our architecture-agnostic proposal. It relies on three key components. We first introduce Functional Instance Normalization layer (Sec.~\ref{sec:fingan-fin}) which enables $\phi$-manifold reshaping. Second, our Disentanglement Residual Block (Sec.~\ref{sec:fingan-DRB}) in charge of $\phi$ disentanglement in input data. Finally, we detail $\phi\text{-Net}$, a pair-wise $\phi$ regression network (Sec.~\ref{sec:fingan-phinet}) which enforces manifold distances consistency.

\paragraph{Model guidance.}
We guide the learning with \textit{simple} non-neural models $M(x,\phi)$, $x$ the source image. Thus, following the intuition that target manifold can be discovered with coarse guidance: night resembles \textit{dark} day, fog looks like a \textit{blurry gray} clear image, etc. We depart from the need of complex physical guidance since we disentangle shared and private features from model/translation which enables discovering complex non-modeled features (e.g. light sources at night). Models are described in Sec.~\ref{sec:exp-methodology} and supp.

\subsection{Functional Instance Normalization (FIN)}
\label{sec:fingan-fin}
To take advantage of our model guidance which is continuous by nature, we must allow our network to encode $\phi$ continuity. 
To do so, we build on prior Instance Normalization (IN) which allows carrying style-related information~\cite{ulyanov2017improved,huang2017arbitrary}. It writes for input $x$,
\begin{equation}
\text{IN}(x)=\frac{x - \mu}{\sigma}\gamma + \beta,
\end{equation}
where $\mu$ and $\sigma$ are input feature statistics, and $\gamma$ and $\beta$ learned parameters of an affine transformation. As an extension, we propose Functional Instance Normalization (FIN)
\begin{equation}
    \text{FIN}(x,\phi)=\frac{x - \mu}{\sigma}f_{\gamma}(\phi) + f_{\beta}(\phi),
\end{equation}
where instead of learning a unique value of affine transformation parameters, 
we learn the distribution of transformations $f_{\gamma}$ and $f_{\beta}$.
The intuition is to shape the $\phi$-manifold based on how the transformation evolves. Compared to others~\cite{gong2019dlow}, this allows us to interpret better the learned manifold.
Depending on the nature of $Y(\phi)$, we can encode FIN layer accordingly. In this work, we investigate linear and cyclic encoding. Linear encoding is commonly encountered, and assumes reorganizing features linearly. For instance, considering adverse weather phenomena, severe conditions (e.g. thick fog) are always positioned after light ones (i.e. lite fog). We model linear FIN parameters as
\begin{equation}
\label{eq:FIN-linear}
\begin{split}
        f_{\gamma}(\phi) = a_{\gamma}\phi + b_{\gamma},\\f_{\beta}\phi = a_{\beta}\phi + b_{\beta},
    \end{split}
\end{equation}
with $\{a_{\gamma}, a_{\beta}, b_{\gamma}, b_{\beta}\}$ the learnable parameters of the layer.

Conversely, some translations path loop back to source, as it happens with daylight, which is \textit{cyclic} by nature going from Day to $\text{Dusk}\mapsto{}\text{Night}\mapsto{}\text{Dawn}$ and Day again. In this case, we encode cyclic FIN layer with parameters
\begin{equation}
\label{eq:FIN-cyclic}
\begin{split}
        f_{\gamma}(\phi) = a_{\gamma}cos(\phi) + b_{\gamma},\\f_{\beta}(\phi) =       a_{\beta}sin(\phi) + b_{\beta}.
\end{split}
\end{equation}

\subsection{Disentanglement Residual Block (DRB)}
\label{sec:fingan-DRB}
The pitfall of strict model-dependency is that the GAN will only learn to mimic the model.
To prevent that, we must allow target domain $Y(\phi)$ and \textit{model} domain $Y_M(\phi)$ to have shared \textit{modeled} features $Y^{\phi}$ but also private \textit{non-modeled} features $Y^E$ and $Y_M^E$, respectively. This writes 
\begin{equation}
	\begin{split}
		Y(\phi) = \{Y^{\phi}, Y^E\}\,,\\
		Y_M(\phi) = \{Y^{\phi}, Y_M^E\}\,.\\
	\end{split}
\end{equation}

\noindent{}We enable private features in either domain with our Disentanglement Residual Block (DRB, shown in Fig.~\ref{fig:method}) whose goal is to extract disentangled representations for a given~$\phi$. The DRB is composed of residual blocks mapping the encoder feature map $h^X$ to the disentangled representations of output images. Let $y^{\phi} \in Y(\phi), y^{\phi}_M \in Y_M(\phi),$ we have 
\begin{equation}
\begin{split}
    \text{DRB}(h^X, \phi) = \{h^Y, h^Y_M\},\\
    y^{\phi} = G_D(h^Y)\,,\;\;\;\;\;\; y^{\phi}_M = G_D(h^Y_M).\\
\end{split}
\end{equation}
The DRB works as follows. 
Following Fig.~\ref{fig:method}, the input representation $h^X$ is processed by residual blocks, each one extracting features associated with the atomic ones previously introduced, such as $Y^{\phi}, Y^E,Y_M^E \longleftrightarrow h^{\phi}, h^E,h_M^E$, one per residual. 
In particular, the residual block for $h^{\phi}$ extraction uses our FIN layers for normalization to encode continuous features. The hidden latent representations $h^Y$ and $h^Y_M$ are obtained from summation of the disentangled features and $h^X$ to ease gradient propagation as in~\cite{he2016deep}. In formulas,
\begin{equation}
\begin{split}
    h^Y &= h^{\phi} + h^E + h^X\,,\\
    h^Y_M &= h^{\phi} + h^E_M + h^X\,.\\
\end{split}
\end{equation}
Intuitively, for optimization we need feedback from both real data similarity and mimicking of the model output. 
While the first must rely on adversarial training due to the use of unpaired images, we can enforce reconstruction on the paired modeled $\tilde{y}^{\phi}_M = M(x, \phi)$. Assuming LSGAN~\cite{mao2017least} training and discriminator $D$, we obtain
\begin{equation}
\begin{split}
\mathcal{L}_{adv}^G &= ||D(y^{\phi}) - 1||_2,\\
\mathcal{L}_{M} &= ||y_M^{\phi} - \tilde{y}_M^{\phi}||_1.\\
\end{split}
\end{equation}
Minimization of $\mathcal{L}_{adv}^G$ and $\mathcal{L}_{M}$ during the generator update step enables disentanglement of $h^E$ and $h^E_M$.

\subsection{Pairwise regression network ($\phi\text{-Net}$)}
\label{sec:fingan-phinet}
The DRB enforces both disentanglement and manifold shape at a feature level, but it requires ad-hoc training strategies to actually disentangle also continuous features for real images and not fall into easy pitfalls, e.g. the network only exploiting $h^E$ for target translation ignoring $h^{\phi}$. Hence, we introduce a training strategy based on similarities which forces the network to both exploit extracted continuous information and follow the model guidance.
Suppose an input image $x$, mapped to $x\mapsto y^{\phi}$ by the network. As shown in Fig.~\ref{fig:method}, we randomly sample $\phi$ and $\phi'$ and apply $M(.)$ to $x$, obtaining the couple $\{\tilde{y}_M^{\phi},\tilde{y}_M^{\phi'}\}$.
We use a CNN ($\phi\text{-Net}$) for domain similarity discovery. It takes as input a pair of images and regresses their $\phi$ differences, such as
\begin{equation}
        \phi\text{-Net}(y^{\phi},y^{\phi'}) = \phi - \phi' = \Delta \phi\,.
\end{equation}
We jointly optimize $\phi\text{-Net}$ and generator ($G$) parameters in an end-to-end setting by enforcing consistency between real and modeled target domain images. In formulas,
\begin{gather}
\begin{aligned}
    \mathcal{L}^G_{\phi} &= ||\phi\text{-Net}(y^{\phi},\tilde{y}_M^{\phi})||_2 + ||\phi\text{-Net}(y^{\phi},\tilde{y}_M^{\phi'}) - \Delta \phi||_2, \\
    \mathcal{L}_{gt} &= ||\phi\text{-Net}(\tilde{y}_M^{\phi}, \tilde{y}_M^{\phi'}) - \Delta \phi||_2, \\
    \mathcal{L}_{\phi} &= \mathcal{L}^G_{\phi} + \mathcal{L}_{gt}.\raisetag{1\baselineskip}
\end{aligned}
\end{gather}
$\mathcal{L}^G_{\phi}$ forces $G$ to organize the manifold following the feedback of the physical model, ultimately resulting in generated $y^{\phi}$ and $\tilde{y}^{\phi}_M$ to be mapped to the same $\phi$ on the manifold discovered by $\phi\text{-Net}$. 
That way, the network can identify that images follow some similarity criteria despite differences between model output and learned translation, {leading to an organization of the latent space guided by the physical model}. 
$\mathcal{L}_{gt}$ instead exploits modeled data only and thus is used to avoid training collapse. For linear FIN, we train on $\phi$ and $\Delta \phi$, though for cyclic one stability is increased by evaluating each loss on sin/cos projection of~$\phi$.

\subsection{Training strategy}
\label{sec:comogan-training}
CoMoGAN is end-to-end trainable and can be used with any i2i framework by simply adding the DRB  between encoder and decoder, with our losses. The final objective for the generator depends if source and target are detached, i.e. $X\not\subset{}Y{}$~(see Fig.~\ref{fig:teaser} for visualization). If detached, the generator update step writes
\begin{equation}
    \mathcal{L}^G = \mathcal{L}^G_{adv} + \mathcal{L}_M + \mathcal{L}_{\phi}\,.
\end{equation}
For attached source/target, we enforce source ($\phi_0$) identity:
\begin{equation}
	\mathcal{L}^G = \mathcal{L}^G_{adv} + \mathcal{L}_M + \mathcal{L}_{\phi} + ||G(x, \phi_0) - x||_1\,.
\end{equation}
Either $\mathcal{L}^G$ definition is used, sometimes in conjunction with a regularization pairwise loss to ease training (cf. supp).\\
Using real data ($\tilde{y}$) from target the discriminator minimizes 
$$ \mathcal{L}^D = \mathcal{L}_{adv}^D = ||D(y^{\phi})||_2 + ||D(\tilde{y}) - 1||_2\,.$$

\begin{figure}
	\centering
	\includegraphics[width=\linewidth]{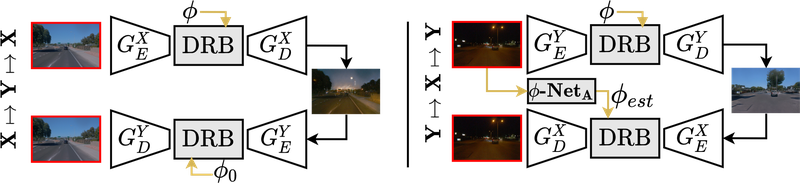}
	\caption{We enforce cycle consistency by injecting the source $\phi_0$ in the $X\mapsto Y \mapsto X$ translation when reconstructing the original image. Also, for $Y\mapsto X \mapsto Y$ we position the input image at $\phi_{est}$ on the domain using our $\phi\text{-Net}_A$ CNN trained unsupervised for $\phi$ regression.}
	\label{fig:implementation}
\end{figure}

\paragraph{Cycle consistency.}\label{method:implementation}
In addition to $X\mapsto Y$, many networks perform $Y \mapsto X$ to preserve context with cycle consistency. 
To handle the latter, we insert a \textit{shared} DRB between each encoder/decoder couple to benefit from multiple sources. This is illustrated in Fig.~\ref{fig:implementation}. 
We also use another unsupervised network, called $\phi\text{-Net}_A$, that regresses $\phi$ on the target dataset.
From above figure (left), because $\phi$ is injected in $X\mapsto{}Y$ transformation, we enforce a correct spreading of all $\phi$ values by adding $\mathcal{L}_{reg}$ to the generator objective, $\mathcal{L}_{reg} = ||\phi\text{-Net}_A(y^{\phi}) - \phi||_2$.

\section{Experiments}
\label{sec:exp}
We show the efficiency of CoMoGAN on new continuous image-to-image translation tasks $X\mapsto{}Y(\phi)$, where we consider source data to lie on a fixed point ($\phi_0$) of the $\phi$-manifold and \textit{unknown} $\phi$ target data. The underlying optimization challenge is to learn simultaneously the $\phi$-manifold and continuous image translation. Because continuous model-guided translation is new, we first describe our three novel translations tasks~(Sec.~\ref{sec:exp-methodology}) obtained by leveraging recent datasets~\cite{sun2020scalability,ros2016synthia,cordts2016cityscapes,halder2019physics,zhu2017unpaired}. Each task encompasses challenges of its own such as linear/cyclic target manifold, attached/detached manifolds (i.e. $X\subset{}Y{}$ or $X\not\subset{}Y$) and uni-/multi- modality. 
Specifically, we train with backbone MUNIT~\cite{huang2018multimodal} (multi-modal) or CycleGAN~\cite{zhu2017unpaired} (uni-modal) and coin our alternatives $\text{CoMo-MUNIT}$ and $\text{CoMo-CycleGAN}$, respectively.
We evaluate the manifold organization~(Sec.~\ref{sec:exp-manifold}) and the translation quality~(Sec.~\ref{sec:exp-translationquali}) from GAN metrics and proxy tasks. Continuous translation (Sec.~\ref{sec:exp-continuous-translation}) is evaluated separately and we conclude with ablation studies~(Sec.~\ref{sec:exp-ablation}). We mostly train with default backbone hyperparameters, more details are in supplementary.

\subsection{Translation tasks}
\label{sec:exp-methodology}
\paragraph{$\text{Day}\mapsto\text{Timelapse}$.} Using recent Waymo Open dataset \cite{sun2020scalability}, we frame the complex task of day to any time, thus learning \textit{timelapse} passing through day/dusk/night/dawn. Waymo image labels are \textit{only} used to split clear images into \textit{source} \{Day\} and \textit{target} \{Dusk/Dawn, Night\}, respectively obtaining train/val sets of 105307 / 28165 and 27272 / 7682 images. 
We train $\text{CoMo-MUNIT}$ for multi-modality. To respect the cyclic nature of time we exploit cyclic FIN (Eq.~\ref{eq:FIN-cyclic}) encoding \corr{sun angles as}{$\phi \in [0, 2\pi]$, which maps to a sun elevation $\in [+30\degree, -40\degree]$}. \textit{For evaluation only}, we obtain ground truth elevation from astronomical models~\cite{pysolar} with image GPS position and timestamp. For guidance, we exploit a simple day-to-night tone mapping~\cite{thompson2002spatial} ($\Omega$) interpolating with Hosek radiance model~\cite{hosek2012analytic} ($\text{HSK}$) to account for gradual loss of color, and adding asymmetrical hue correction ($\text{corr}$) to account for temperature changes -- i.e. at analog sun elevation dusk appears red-ish and dawn purple-ish --. The complete model is in the supplementary. It writes
\begin{equation} 
	\label{eq:exp-models-daytimelapse}
	\begin{split}
		M(x, \phi) &= (1 - \alpha)x+ \alpha \Omega(x, \text{HSK}(\phi) + \text{corr}(\phi)) + \text{corr}(\phi).
	\end{split}
\end{equation}

\paragraph{$\text{iPhone}\mapsto\text{DSLR}$.}
We inspire from CycleGAN~\cite{zhu2017unpaired} by adapting their initial task to a continuous setup, learning the mapping of iPhone images with large depth of field to DSLR images with shallow depth of field. We also use the iphone2dslr flowers dataset~\cite{zhu2017unpaired}, split in \textit{source} 1182/569 and \textit{target} 3325/480. We train this task with $\text{CoMo-CycleGAN}$ for comparison, and use linear FIN~(Eq.~\ref{eq:FIN-linear}) where $\phi \in [0, 1]$ encodes the progression.

For guidance, we naively render blur by convolving ($\ast{}$) a Gaussian ($G$) which kernel size maps to $\phi$. That is
\begin{equation} 
	\label{eq:exp-models-DSLR}
	M(x, \phi) = G(\phi)\ast{}x\,.
\end{equation}

\paragraph{$\text{Synthetic}_{\text{clear}}\mapsto\text{Real}_{\text{clear, foggy}}$.}\label{experiments:setups-synth2real}
Here, we propose a detached source/target task, where we learn clear to foggy except that source is \textit{synthetic} and target is \textit{real} data. 
For \textit{source}, we leverage spring sequences of synthetic Synthia dataset~\cite{ros2016synthia}, split in 3497/959 images.  As \textit{target} we mix original Cityscapes~\cite{cordts2016cityscapes} and 4 augmented foggy Weather~Cityscapes~\cite{halder2019physics} with max visibility distances \{750m, 350m, 150m, 75m\}. In target, each of the 5 Cityscapes version has 2975/500 images. We train here a $\text{CoMo-MUNIT}$ with linear FIN layer (Eq.~\ref{eq:FIN-linear}) and encode maximum visibility as \corr{$\phi \in [\infty,70m]$}{$\phi \in [0, 1]$, i.e. visibility $\in [\infty,70m]$}. For guidance, we simply exploit the fog model of~\cite{halder2019physics}. For the sake of space, models details, sample outputs and model experiments are provided in the supplementary.

\begin{figure}
    \centering
    \includegraphics[width=\linewidth]{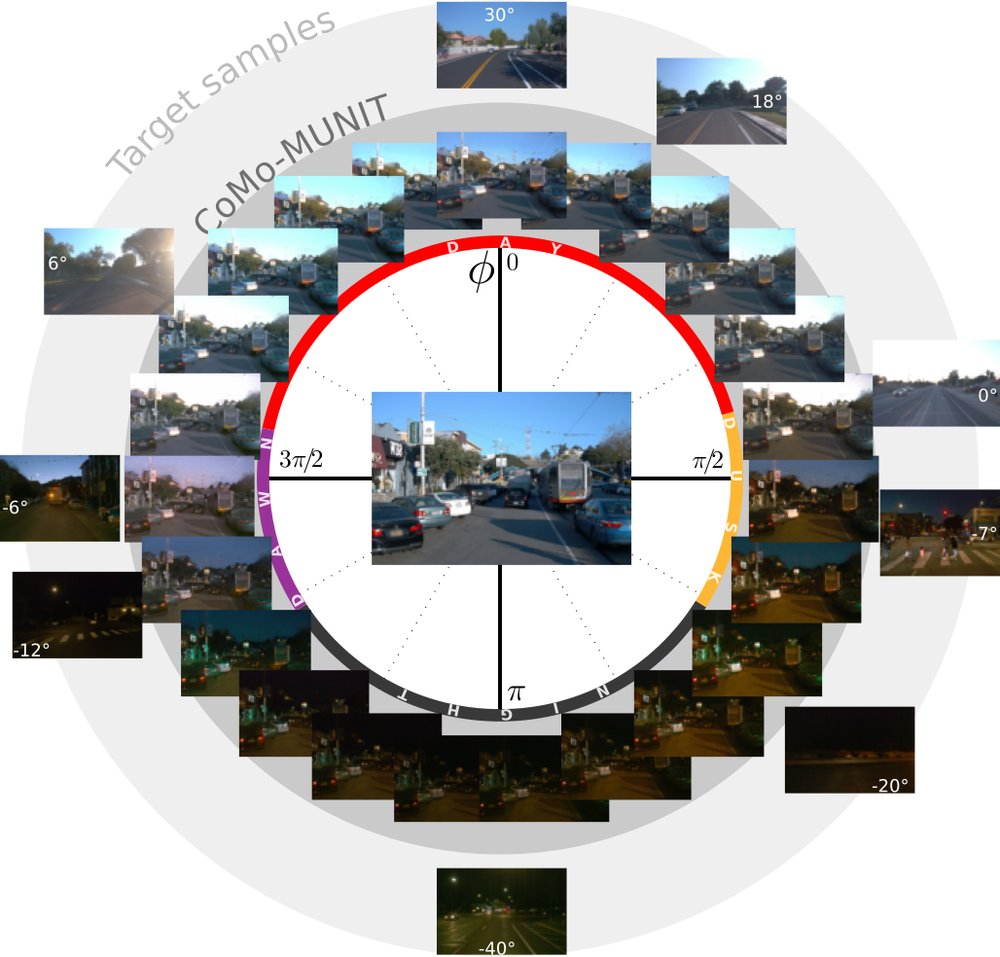}
    \caption{Translations (dark circle) of a source day image (center) exhibit both high variability and similarities with target data (outer circle) for which we report ground truth elevations. CoMo-MUNIT learned non-modeled visual features like frontal sun scenes resembling real ones (as in $\{0\degree, 6\degree, 18\degree\}$). Note that it discovered dawn/dusk and the stationary appearance of night, proving manifold quality.}
    \label{fig:fig_cyclic}\vspace{-0.5em}
\end{figure}

\begin{figure}
	\includegraphics[width=\linewidth]{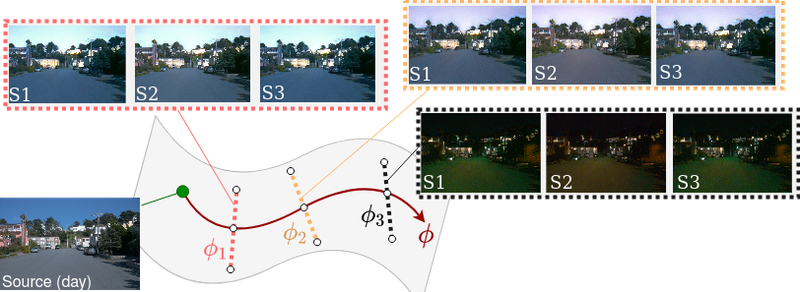}
	\caption{Translations along dimensions $\phi$ (red) and style (dotted). For a given $\phi$, the styles vary slightly (notice hue and brightness), proving disentanglement of $\phi$ and style.}
	\label{fig:multimodal}
\end{figure}

\subsection{Manifold organization}
\label{sec:exp-manifold}
We evaluate the quality of the unsupervised manifold discovery using $\text{CoMo-MUNIT}$ on the $\text{Day}\mapsto\text{Timelapse}$.
Fig.~\ref{fig:fig_cyclic} shows a source day image (center) and our timelapse translations for uniformly sampled $\phi$ (middle circle). 
Apart from the appealing translations appearance, notice the network discovered important features like frontal sun (when the sun is close to the horizon), sunset/sunrise, material reflectance (at night), and the stable nighttime appearance. All these features are not in model $M(.)$ though present in target images (outer circle). 
This advocates the network disentangled model features and translation features. Note also that the top translation \corr{($\phi=30\degree$)}{in Fig.~\ref{fig:fig_cyclic}} accurately resembles source, assessing that target is attached to source.

Quantitatively, we measure the manifold precision by regressing $\phi$ with our $\phi\text{-Net}_A$ CNN (cf. Sec.~\ref{sec:comogan-training}) on real Waymo validation set, and compute the error w.r.t. ground truth \corr{sun angles}{elevations}. We get a mean error of \corr{$12.59\degree$~(std~$6.73\degree$)}{$19.8\degree$~(std~8.56$\degree$)} when \textit{unsupervised} and $4.05\degree$~(std $4.20\degree$) if \textit{supervised}.
Even unsupervised, our manifold discovery is acceptable, and opens ways for unsupervised translations where $\phi$ ground truth would be impractical (e.g. rain, snow).

\paragraph{Disentangled dimensions.}
Because MUNIT is multi-modal by design, it is important to assess $\text{CoMo-MUNIT}$ properly disentangles $\phi$ from the style dimension of MUNIT. We do this by sampling $\phi$ and style. From Fig.~\ref{fig:multimodal}, the latter evolve correctly on different axes, which was expected since $\phi$ is regulated by model-guided features. 
Again, using $\phi\text{-Net}_A$, we regress $\phi$ values for 100 fixed $\phi$ translations each with 100 different styles, obtaining \corr{$0.74\degree$}{$1.06\degree$} $\phi$-variance along the style dimension. This proves the orthogonality of $\phi$ and style manifolds.

\subsection{Translation quality}
\label{sec:exp-translationquali}

\paragraph{GAN metrics.}
We measure the quality and variability of all translations task w.r.t. MUNIT and CycleGAN backbones, showcasing in Tab.~\ref{tab:gan-metrics} that we always perform better or on par. In the table, IS~\cite{salimans2016improved} evaluates image quality and diversity over all the dataset, CIS~\cite{huang2018multimodal} over multimodal translations, and LPIPS~\cite{zhang2018unreasonable} evaluates absolute diversity only. 
We conjecture our performance results of the higher degree of control we have, since we control $\phi$ features in a disentangled manner (i.e. extremely increasing variability), while entangled backbones lean towards the easiest translations. The InceptionV3 networks used for IS/CIS evaluation are trained on the source/target classification task. IS is evaluated on all validation set, while for CIS/LPIPS we follow~\cite{huang2018multimodal} evaluation routine.

\begin{table}
        \centering
        \scriptsize
    	\setlength{\tabcolsep}{0.007\linewidth}
            \begin{tabular}{ccccc}
            \toprule
            \textbf{Task}                  & \textbf{Network}          & \textbf{IS$\uparrow$}    & \textbf{CIS$\uparrow$} & \textbf{LPIPS$\uparrow$}   \\\hline
            \multirow{2}{*}{$\text{Day}\mapsto\text{Timelapse}$}      & MUNIT~\cite{huang2018multimodal} & 1.43 & 1.41 & \textbf{0.583} \\
           & $\text{CoMo-MUNIT}$             & \corr{}{\textbf{1.59}} & \corr{}{\textbf{1.51}} & \corr{}{0.580} \\ \hline %
            \multirow{2}{*}{$\text{Syn.}_\text{clear}\mapsto\text{Real}_\text{clear, foggy}$}     & MUNIT~\cite{huang2018multimodal} & \textbf{1.30} & 1.02 & 0.493 \\
                                       & $\text{CoMo-MUNIT}$           & \textbf{1.30} & \textbf{1.05} & \textbf{0.515} \\\hline
            \multirow{2}{*}{$\text{iPhone}\mapsto\text{DSLR}$} & CycleGAN~\cite{zhu2017unpaired} & 1.39 & n.a.* & 0.658 \\
                                       & $\text{CoMo-CycleGAN}$            & \textbf{1.44} & \textbf{1.18} & \textbf{0.680}\\
            \bottomrule
        
            \end{tabular}\\
        * CIS is only applicable to multi-modal network.  
        \caption{GAN metrics proves the benefit of our controllable $\phi$ generation, leading to on par or better quality/variability.}
        \label{tab:gan-metrics}
\end{table}\vspace{-1em}
\begin{figure}
	\begin{subfigure}[t]{0.36\linewidth}
	    \scriptsize
		\setlength{\tabcolsep}{0.01\linewidth}
	        \begin{tabular}{ccc|c}
	        \toprule
			\textbf{Translations} & \textbf{CS} & \textbf{FD} & \textbf{Mean} \\
	        \midrule
	        none (source) & 10.9 & 10.1 & 10.5\\
	        Model~\cite{halder2019physics} & 19.9 & 21.5 & 20.7\\
	        MUNIT~\cite{huang2018multimodal} & 38.3 & 21.8 & 30.0\\
	        \midrule
	        $\text{CoMo-MUNIT}$ & \textbf{43.0} & \textbf{23.4} & \textbf{33.2}\\
	        \bottomrule
	        \end{tabular}
	\caption{mIoU metric}
	\label{tab:segmentation-miou}
	\end{subfigure}\hspace{0.01\linewidth}\begin{subfigure}[t]{0.63\linewidth}
		\resizebox{\linewidth}{!}{
		\tiny
		\setlength{\tabcolsep}{0.003\linewidth}
		\begin{tabular}{c c c c}
			\includegraphics[width=4em, valign=m]{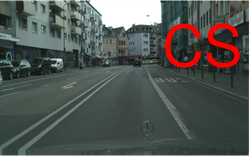}
			& \includegraphics[width=4em, valign=m]{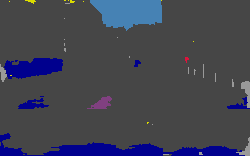}
			& \includegraphics[width=4em, valign=m]{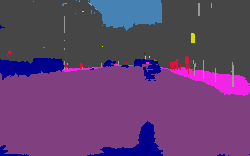}
			& \includegraphics[width=4em, valign=m]{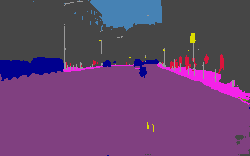}
			\\
			\includegraphics[width=4em, valign=m]{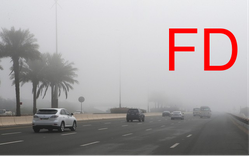}
			& \includegraphics[width=4em, valign=m]{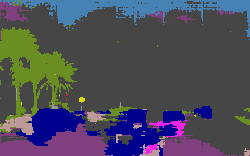}
			& \includegraphics[width=4em, valign=m]{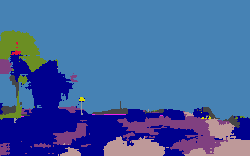}
			& \includegraphics[width=4em, valign=m]{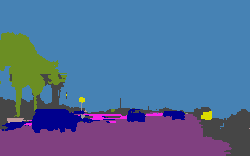}
			\\
			Input & Model & MUNIT & Ours
		\end{tabular}}
	\caption{Samples}
	\label{fig:qualitative-segmentation}
	\end{subfigure}
\caption{Semantic segmentation on clear Cityscapes~(CS)~\cite{cordts2016cityscapes} and Foggy Driving (FD)~\cite{sakaridis2018semantic} with PSPNet-50~\cite{zhao2017pyramid} trained on clear Synthia (\textit{source}), foggy physics \textit{Model}, and $\text{Synthetic}_{\text{clear}}\mapsto\text{Real}_{\text{clear, foggy}}$ of MUNIT or CoMo-MUNIT. Noticeably, we outperform all on both clear (CS) and foggy (FD) dataset.}\label{fig:quali-quanti-fog}\vspace{-1em}
\end{figure}

\paragraph{Semantic segmentation.} 
We measure the effectiveness of our $\text{Synthetic}_{\text{clear}}\mapsto\text{Real}_{\text{clear, foggy}}$ translations in Fig.~\ref{fig:quali-quanti-fog} by training PSPNet-50~\cite{zhao2017pyramid} with either MUNIT or CoMo-MUNIT outputs. 
For comparison, we also train segmentation with clear \textit{source} Synthia or physics-based foggy \textit{model}~\cite{halder2019physics} as for guidance.
For MUNIT and CoMo-MUNIT, we employ a multi-modal style-sampling strategy~\cite{pizzati2019domain} with 5 fixed styles. Additionally, for CoMo-MUNIT and \textit{model} translations that allow it, we sample uniform $\phi$. We follow \cite{zhao2017pyramid} settings and train 150 epochs, using 3498 train images for each setup.

Tab.~\ref{tab:segmentation-miou} reports the standard mIoU on shared Synthia-Cityscapes classes on real images from the validation set of Cityscapes~\cite{cordts2016cityscapes} (CS, 500 images) and Foggy Driving~\cite{sakaridis2018semantic} (FD, 101 images). While the transformation is subtle, it still reduces the domain shift, since even if \textit{Model} significantly outperforms \textit{source} but we beat all by additional margin of +4.7/+1.6/+3.2. Noticeably, we improve both on clear (CS) and foggy (FD) datasets showing CoMo-MUNIT preserved accurate clear \textit{and} foggy translations. We speculate instead that MUNIT focuses on target dataset fog intensities which are discrete and may differ from FD, while our FIN layer enables continuous representation leading to better generalization.
Qualitative evaluation on both datasets in Fig.~\ref{fig:qualitative-segmentation} respects mIoU performances.

\begin{figure*}
	\centering
	\resizebox{\linewidth}{!}{
	\setlength{\tabcolsep}{0.003\linewidth}
	\tiny
	\begin{tabular}{c c c c c c c c c c c}
			\multicolumn{11}{c}{\textbf{Multi-target i2i}}\\
			\multirow{1}{*}[0cm]{\rotatebox{90}{\cite{choi2020stargan}}} & \multirow{1}{*}[0.35cm]{\rotatebox{90}{StarGAN V2}}
            & \includegraphics[width=10em, valign=m]{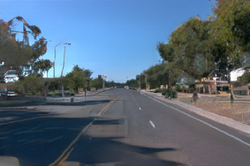}
            & \includegraphics[width=10em, valign=m]{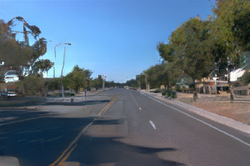}
            & \includegraphics[width=10em, valign=m]{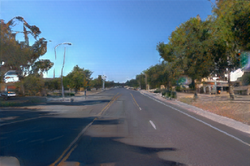}
            & \includegraphics[width=10em, valign=m]{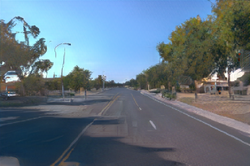}
            & \includegraphics[width=10em, valign=m]{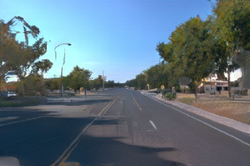}
            & \includegraphics[width=10em, valign=m]{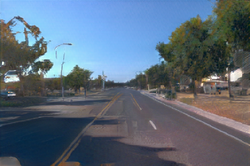}
            & \includegraphics[width=10em, valign=m]{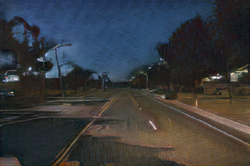}
            & \includegraphics[width=10em, valign=m]{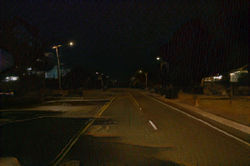}
            & \includegraphics[width=10em, valign=m]{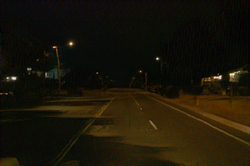}
            \\
			\multicolumn{11}{c}{\textbf{Continuous linear i2i}}\\
			\multirow{1}{*}[0.15cm]{\rotatebox{90}{\cite{gong2019dlow}}} & \multirow{1}{*}[0.2cm]{\rotatebox{90}{DLOW}}
			& \includegraphics[width=10em, valign=m]{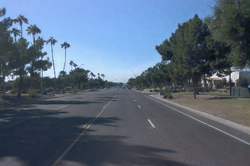}
			& \includegraphics[width=10em, valign=m]{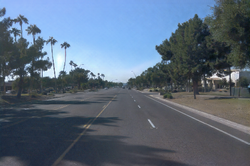}
			& \includegraphics[width=10em, valign=m]{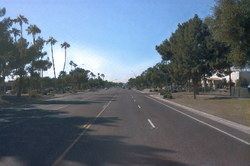}
			& \includegraphics[width=10em, valign=m]{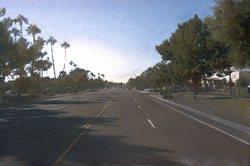}
			& \includegraphics[width=10em, valign=m]{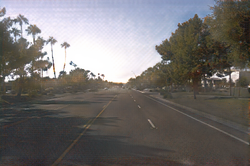}
			& \includegraphics[width=10em, valign=m]{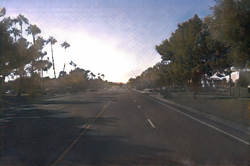}
			& \includegraphics[width=10em, valign=m]{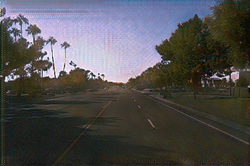}
			& \includegraphics[width=10em, valign=m]{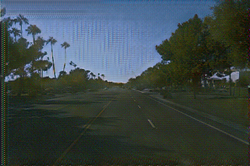}
			& \includegraphics[width=10em, valign=m]{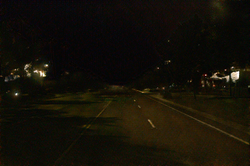}
			\\
			
			\multirow{1}{*}[0.15cm]{\rotatebox{90}{\cite{wang2019deep}}} & \multirow{1}{*}[0.53cm]{\rotatebox{90}{DNI-CycleGAN}}
			& \includegraphics[width=10em, valign=m]{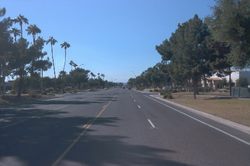}
			& \includegraphics[width=10em, valign=m]{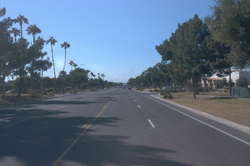}
			& \includegraphics[width=10em, valign=m]{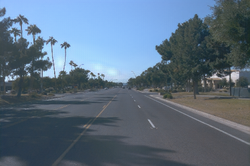}
			& \includegraphics[width=10em, valign=m]{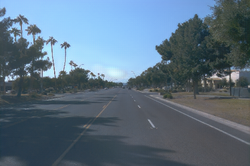}
			& \includegraphics[width=10em, valign=m]{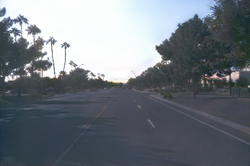}
			& \includegraphics[width=10em, valign=m]{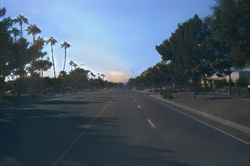}
			& \includegraphics[width=10em, valign=m]{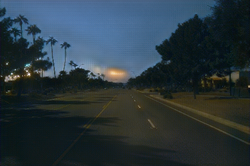}
			& \includegraphics[width=10em, valign=m]{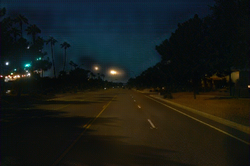}
			& \includegraphics[width=10em, valign=m]{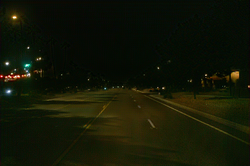}
			\\
			\multirow{1}{*}[0.15cm]{\rotatebox{90}{\cite{wang2019deep}}} & \multirow{1}{*}[0.35cm]{\rotatebox{90}{DNI-MUNIT}}
			& \includegraphics[width=10em, valign=m]{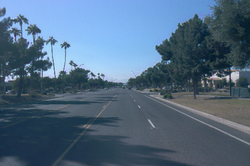}
			& \includegraphics[width=10em, valign=m]{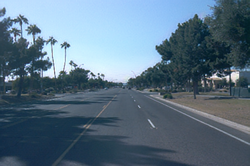}
			& \includegraphics[width=10em, valign=m]{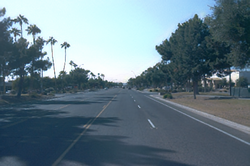}
			& \includegraphics[width=10em, valign=m]{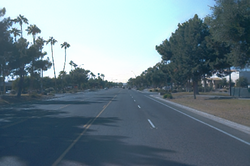}
			& \includegraphics[width=10em, valign=m]{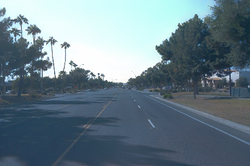}
			& \includegraphics[width=10em, valign=m]{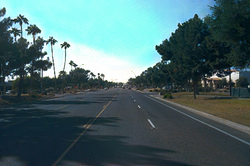}
			& \includegraphics[width=10em, valign=m]{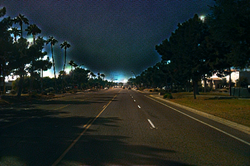}
			& \includegraphics[width=10em, valign=m]{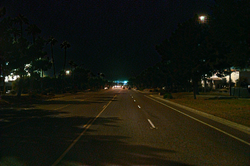}
			& \includegraphics[width=10em, valign=m]{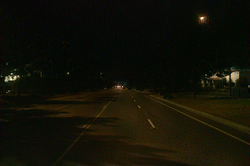}
			\\
			\midrule
			\multicolumn{11}{c}{\textbf{Continuous cyclic i2i}}\\
			& \multirow{2}{*}[0cm]{\rotatebox{90}{\textbf{CoMo-MUNIT}}}
            & \multirow{2}{*}[2.5em]{\includegraphics[width=10em]{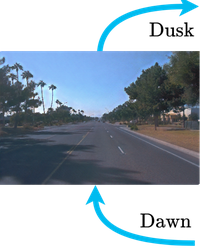}}
            & \includegraphics[width=10em, valign=m]{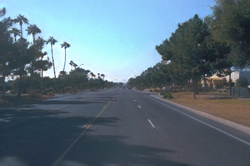}
			& \includegraphics[width=10em, valign=m]{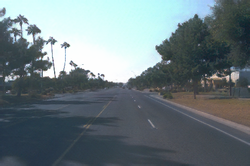}
			& \includegraphics[width=10em, valign=m]{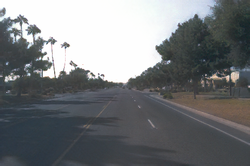}
			& \includegraphics[width=10em, valign=m]{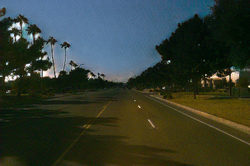}
			& \includegraphics[width=10em, valign=m]{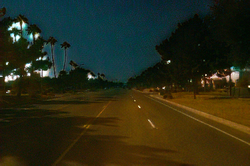}
			& \includegraphics[width=10em, valign=m]{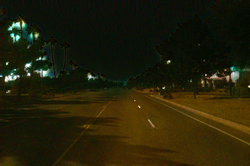}
			& \includegraphics[width=10em, valign=m]{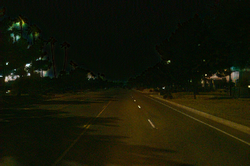}
            & \multirow{2}{*}[2.5em]{\includegraphics[width=10em, valign=m]{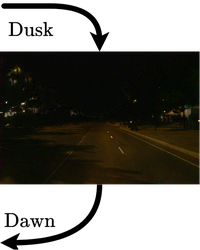}}
			\\
			& &
			& \includegraphics[width=10em, valign=m]{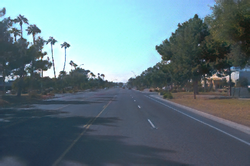}
			& \includegraphics[width=10em, valign=m]{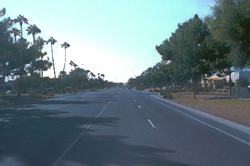}
			& \includegraphics[width=10em, valign=m]{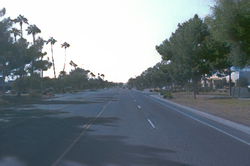}
			& \includegraphics[width=10em, valign=m]{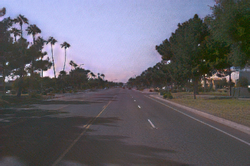}
			& \includegraphics[width=10em, valign=m]{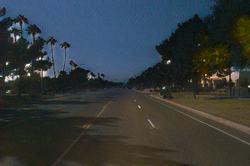}
			& \includegraphics[width=10em, valign=m]{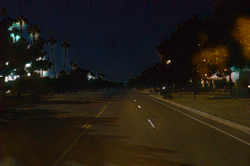}
			& \includegraphics[width=10em, valign=m]{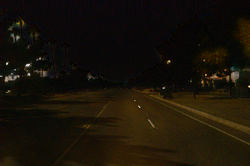}
			&
			\\

			\bottomrule
				        \multicolumn{11}{l}{
	        \begin{tikzpicture}
        
            \pgfdeclarepatternformonly{senwStripes}{\pgfpoint{0cm}{0cm}}{\pgfpoint{1cm}{1cm}}{\pgfpoint{1cm}{1cm}}
            {
                \foreach \i in {0.0cm, 0.2cm,...,0.8cm}
                {
                 \pgfpathmoveto{\pgfpoint{\i}{0cm}}
                 \pgfpathlineto{\pgfpoint{0cm}{\i}}
                 \pgfpathlineto{\pgfpoint{0cm}{\i + 0.1cm}}
                 \pgfpathlineto{\pgfpoint{\i + 0.1cm}{0cm}}
                 \pgfpathclose%
                 \pgfusepath{fill}
                 \pgfpathmoveto{\pgfpoint{1cm}{\i}}
                 \pgfpathlineto{\pgfpoint{\i}{1cm}}
                 \pgfpathlineto{\pgfpoint{\i + 0.1cm}{1cm}}
                 \pgfpathlineto{\pgfpoint{1cm}{\i + 0.1cm}}
                 \pgfpathclose%
                 \pgfusepath{fill}
                }
            }

            \fill[fill=cyan,cyan] (0,0) rectangle (7.37, 0.2) node[pos=.5,white,font=\bfseries] {Day};
            \fill[preaction={fill={rgb,255:red,152; green,49; blue,152}},pattern=senwStripes, pattern color={rgb,255:red,255; green,182; blue,50}]  (7.37, 0) rectangle (11.79, 0.2)node[pos=.5,white,font=\bfseries] {Dawn/Dusk};
            \fill[fill=black,black]  (11.79, 0) rectangle (17.2, 0.2) node[pos=.5,white,font=\bfseries] {Night};
	        \end{tikzpicture}
	        
	        }\\

			&
			& $+30.00\degree$ 
			& $+21.25\degree$ 
			& $+12.50\degree$ 
			& $3.75\degree$ 
			& $-5.00\degree$ 
			& $-13.75\degree$ 
			& $-22.50\degree$ 
			& $-31.25\degree$
			& $-40.00\degree$ 
			\vspace{-1em}
			\\
			\end{tabular}}
		\caption{$\text{Day}\mapsto\text{Timelapse}$ translations. Baselines output unrealistic translations (e.g. DLOW~\cite{gong2019dlow}) or images with limited variability (StarGAN~V2~\cite{choi2020stargan}). DNI~\cite{wang2019deep} is the best baseline, though our CoMo-MUNIT (last row) is the only cyclic one, outputs more variable images (e.g. at dusk/dawn) and discovered stable night \textit{with less supervision}.}\label{fig:qualitative}\vspace{-1em}
\end{figure*}
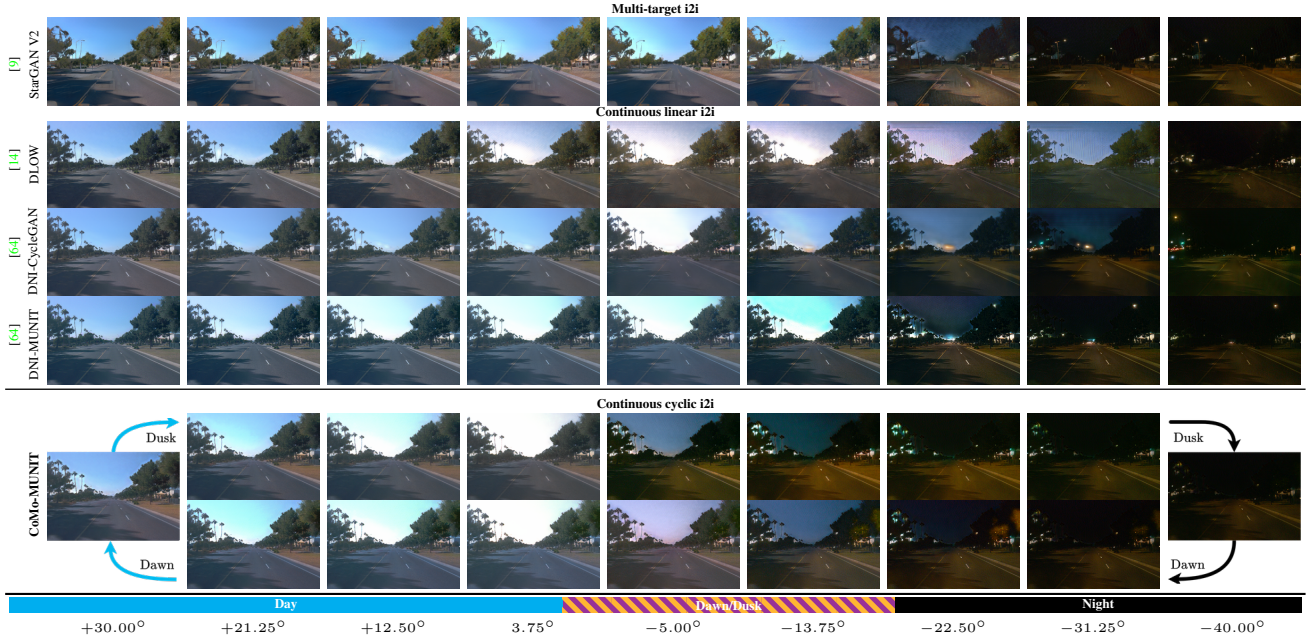

\subsection{Continuous translation quality}
\label{sec:exp-continuous-translation}
\begin{figure}
	\centering
	\resizebox{\linewidth}{!}{
	\setlength{\tabcolsep}{0.003\linewidth}
	\tiny
	\begin{tabular}{c | c c c}
            \includegraphics[width=10em, valign=m]{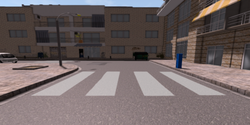}

            & \includegraphics[width=10em, valign=m]{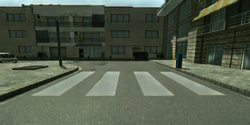}
			& \includegraphics[width=10em, valign=m]{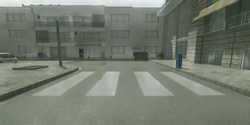}
			& \includegraphics[width=10em, valign=m]{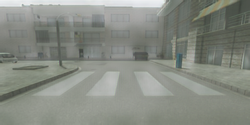}
			\\
            \includegraphics[width=10em, valign=m]{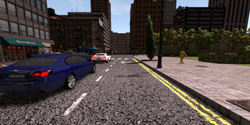}
            & \includegraphics[width=10em, valign=m]{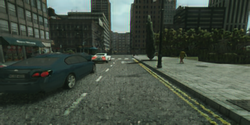}
			& \includegraphics[width=10em, valign=m]{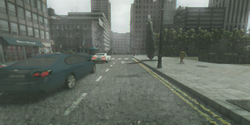}
			& \includegraphics[width=10em, valign=m]{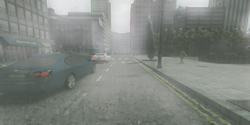}
			\\
            \includegraphics[width=10em, valign=m]{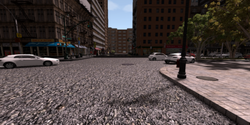}
            & \includegraphics[width=10em, valign=m]{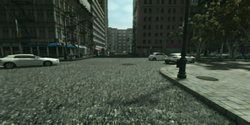}
			& \includegraphics[width=10em, valign=m]{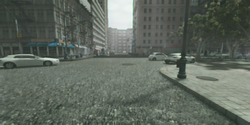}
			& \includegraphics[width=10em, valign=m]{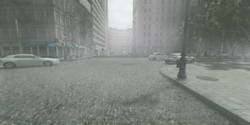}
			\\
			Source &\tikzmark{a}{} &  & \tikzmark{b}{}\\
			Synthetic (clear) & Real (clear) & \textbf{CoMo-MUNIT} & Real (foggy)
        \end{tabular}\link{a}{b}}

		\caption{Sample $\text{Synthetic}_{\text{clear}}\mapsto\text{Real}_{\text{clear, foggy}}$ translations with CoMo-MUNIT. Note the complex detached source (Synthia~\cite{ros2016synthia}) and target (clear/foggy Cityscapes\cite{cordts2016cityscapes,halder2019physics}) setting. Still, clear translations correctly encompass Cityscapes stylistic appearance (notice texture and color).}\label{fig:qualitative-fog}
\end{figure}

\begin{figure}
	\centering
	\resizebox{0.9\linewidth}{!}{
	\setlength{\tabcolsep}{0.01\linewidth}
	\begin{tabular}{c | c c c | c}
            \includegraphics[width=7em, valign=m]{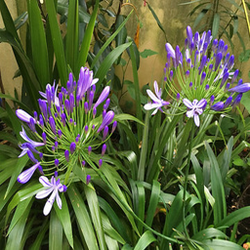}
            & \includegraphics[width=7em, valign=m]{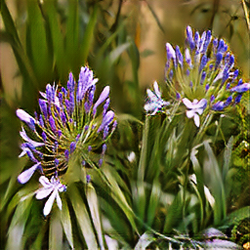}
            & \includegraphics[width=7em, valign=m]{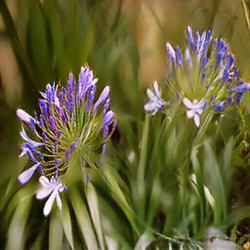}
            & \includegraphics[width=7em, valign=m]{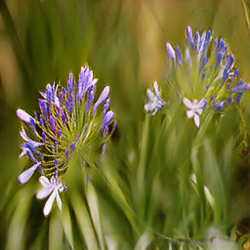}
            & \includegraphics[width=7em, valign=m]{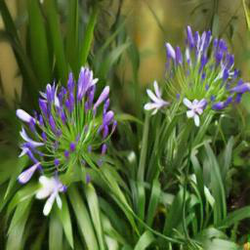}
			\\
            \includegraphics[width=7em, valign=m]{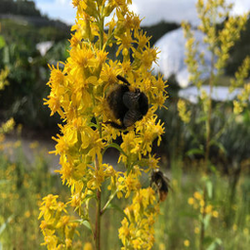}
            & \includegraphics[width=7em, valign=m]{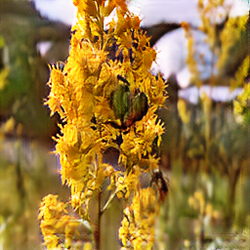}
            & \includegraphics[width=7em, valign=m]{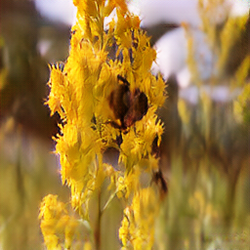}
            & \includegraphics[width=7em, valign=m]{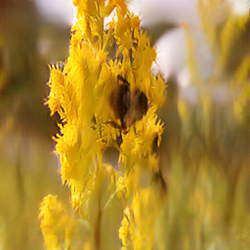}
            & \includegraphics[width=7em, valign=m]{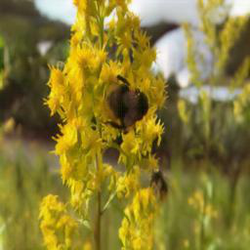}
			\\
            \includegraphics[width=7em, valign=m]{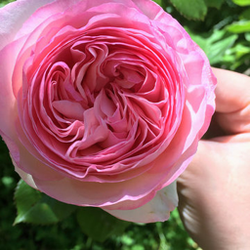}
            & \includegraphics[width=7em, valign=m]{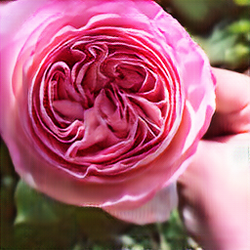}
            & \includegraphics[width=7em, valign=m]{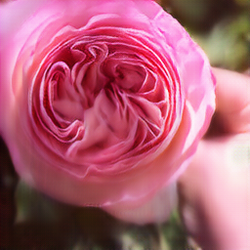}
            & \includegraphics[width=7em, valign=m]{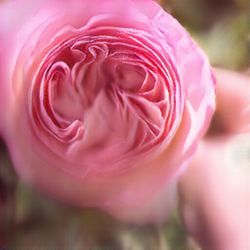}
            & \includegraphics[width=7em, valign=m]{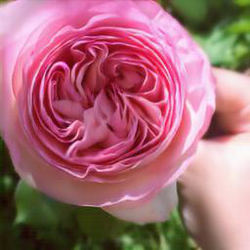}
			\\
            Source & \tikzmark{c}{} & & \tikzmark{d}{} &\\
			iPhone & iPhone & \textbf{CoMo-CycleGAN} & DSLR & CycleGAN~\cite{zhu2017unpaired} \\
        \end{tabular}\link{c}{d}}
		\caption{$\text{CoMo-CycleGAN}$ translations on the  $\text{{iPhone}}\mapsto{}\text{{DSLR}}$ task, using iphone2dslr dataset~\cite{zhu2017unpaired}. Despite naive blur guidance (Eq.~\ref{eq:exp-models-DSLR}), it learns continuous DSLR depth of field, while~\cite{zhu2017unpaired} outputs only target translations.}\label{fig:qualitative-dslr}
\end{figure}

To evaluate the continuity of the translations, we show uniformly spaced $\phi$ translations for $\text{Day}\mapsto\text{Timelapse}$ (Fig.~\ref{fig:qualitative}, bottom row), $\text{Synthetic}\mapsto\text{Real}$ (Fig.~\ref{fig:qualitative-fog}) and $\text{iPhone}\mapsto\text{DSLR}$ (Fig.~\ref{fig:qualitative-dslr}). 
For all, regardless of the backbone and task, our translations look appealing with our network discovering unique visual features \textit{not} present in the model guidance. This is quite noticeable in DSLR (Fig.~\ref{fig:qualitative-dslr}) which learned depth of field despite simple blurring guidance, or in the detached foggy experiment (Fig.~\ref{fig:qualitative-fog}) since translations encompass the desired real appearance with increasing fog.


\subsubsection{Benchmark evaluation} 
We \corr{wish now to }{}evaluate the challenging $\text{Day}\mapsto\text{Timelapse}$ with the literature. 
This is not trivial since our proposal is to the best of our knowledge the first continuous cyclic GAN. 
While \textit{some} previous works could be adapted to cyclic translation (e.g. DLOW~\cite{gong2019dlow}) they \textit{all} require intermediate labeled target points. 
Hence, to achieve a fair comparison compensating data scarcity in Waymo Open, we formulate timelapse as linear \{Day, Dusk/Dawn, Night\} for all baselines and randomly sample between Dusk or Dawn branch with our cyclic network.
Please bear in mind that \textbf{all baselines are more supervised than ours} since they use \corr{the }{}intermediate Dusk/Dawn point while CoMoGAN discovers the manifold from unsupervised target data.  We now \corr{describe all 4 }{detail the }baselines.

\noindent{}\textbf{StarGAN v2}~\cite{choi2020stargan} is a state-of-the-art multi-target i2i architecture learning multiple mapping from the same source point. We train it with official implementation on $\text{Day}\mapsto\text{Dawn/Dusk}\mapsto\text{Night}$ path and use its style code disentanglement capability to enable continuous i2i.\\
\textbf{DLOW}~\cite{gong2019dlow} is continuous by design. We train it with 2 unimodal DLOW $\text{Day}\mapsto\text{Dawn/Dusk}$ and $\text{Dawn/Dusk}\mapsto\text{Night}$. Note that it can be multi-target, but we already compare with the more recent StarGAN v2.\\
\textbf{DNI}~\cite{wang2019deep} applies Deep Network Interpolation to interpolate among kernels of finetuned networks for continuous i2i. We adapt 2 baselines DNI-CycleGAN and DNI-MUNIT both trained on $\text{Day}\mapsto\text{Dawn/Dusk}\mapsto\text{Night}$.\\

\paragraph{Comparison.}
From Fig.~\ref{fig:qualitative}, baselines (rows 1-4) either exhibit limited variability in interpolated points (StarGAN v2 / DNI) or unrealistic results (\corr{DLOW in night rendering}{e.g. DLOW at night}). 
A key limitation is that they rely on (piece-wise) linear interpolation \corr{which makes}{preventing} \corr{them unable to discover}{them from discovering} the stationary \corr{appearance}{aspect} of night (last 3 \corr{columns}{cols}). Conversely,\corr{note our}{} CoMo-MUNIT \corr{method}{}(bottom row) \corr{outputs translation}{translations are} both realistic and stationary at night.

\begin{figure}
	\centering
	\begin{subfigure}[t]{0.5\linewidth}
		\centering
		\adjustbox{valign=c}{
			\includegraphics[width=\linewidth]{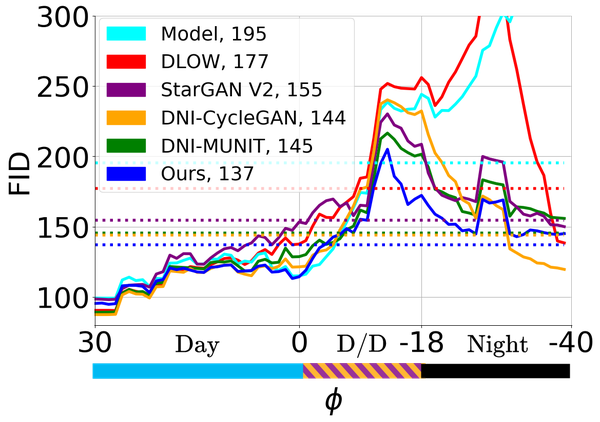}}
		\caption{Rolling FID}\label{fig:fid-plot}
	\end{subfigure}\hspace{0.03\linewidth}\begin{subfigure}[t]{0.47\linewidth}
	    \centering
		\scriptsize
		\setlength{\tabcolsep}{0.002\linewidth}
		\begin{tabular}{ccc}
			\toprule
			\textbf{Method} & \textbf{Mean err. $\downarrow$} & \textbf{Std $\downarrow$} \\
			\midrule
			Model & 21.12 & 10.15 \\
			DLOW~\cite{gong2019dlow} & 17.39 & 9.02 \\ %
			StarGANV2~\cite{choi2020stargan} & 15.91 & 10.00 \\
			DNI-CycleGAN~\cite{wang2019deep} & 13.84 & 7.91 \\
			DNI-MUNIT~\cite{wang2019deep} & 13.80 & 8.30 \\
			\midrule
			CoMo-MUNIT & \corr{}{\textbf{9.84}} & \corr{}{\textbf{7.20}}\\ %
			\midrule
			Real data & 3.61 & 4.52\\
			\bottomrule
		\end{tabular}
		\caption{$\phi$ regression}
		\label{tab:inception_metric}
	\end{subfigure}
	\caption{Evaluation of $\text{Day}\mapsto\text{Timelapse}$. In \protect\subref{fig:fid-plot} rolling FID (cf. text) shows our method is more effective \corr{in complex transformation point like astronomical dawn/dusk ("D/D") and night. This translates as better mean FID (in legend)}{in the complex dawn/dusk ("D/D") and night points, translating as lower mean FID (in legend)}. In \protect\subref{tab:inception_metric}\corr{ we show the mean and std error between input $\phi$ translation values and the regressed $\phi$ with an InceptionV3 network (trained on real data). We rank best on both.}{, we rank best on both mean and std error between the input $\phi$ and the regressed $\phi$ with an InceptionV3 network (trained on real data).}}
	\label{fig:benchmark-perf}
\end{figure}

We also study the realism of all translations using the Frechet Inception Distance~\cite{heusel2017gans} (FID) to measure features distances between generated images and real ones. 
For that, we uniformly split the \corr{sun angles}{elevations} range $[+30\degree, -40\degree]$ in 70 overlapping bins of $7\degree$ width, and compute each bin FID by comparing 100 translations and ad-hoc real images. We refer to this as "rolling FID", plotted in Fig.~\ref{fig:fid-plot}. From the latter, our method outperforms others especially in complex intermediate conditions. Note the baselines performance at precise "dawn/dusk" center (where they are supervised) and how their FID degrade as they depart toward night (approx. $-18\degree$). Even if \textit{unsupervised}, our lower FID shows CoMo-MUNIT better learned these complex visual transitions.\\
An alternative accuracy evaluation is proposed with a proxy task, which is an InceptionV3 network trained to regress sun \corr{angle}{elevation} from real images and $\phi$ ground truths. 
For each method, we then generate 100 images at 100 $\phi$ locations, and measure the error between the input $\phi$ and the inference with the InceptionV3. Tab.~\ref{tab:inception_metric} shows we outperform other methods with a $3.96\degree$ margin due to our better mapping.

\begin{figure}
    \centering
	\scriptsize
	\setlength{\tabcolsep}{0.005\linewidth}
	\def\arraystretch{0.5}%
	\begin{tabular}{ccc}
	    \centering
		\includegraphics[width=0.36\linewidth]{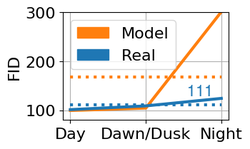}
		& \includegraphics[width=0.29\linewidth]{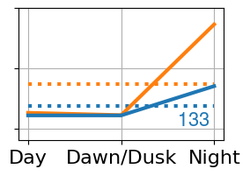}
		& \includegraphics[width=0.29\linewidth]{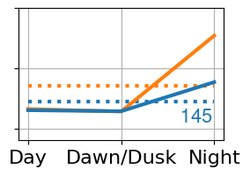}\hfill\\
		(a) Ours
		& (b) \cite{jahanian2019steerability}~($\lambda = 1$)
		& (c) \cite{jahanian2019steerability}~($\lambda = 5$)
	\end{tabular}
	\caption{FIDs (cf. text) for \textit{ours} (a) and steerable GANs~\cite{jahanian2019steerability} (b-c). 
	\corr{Note \textit{ours} has lowest FIDs, because it learned both shared real/model features and private ones.}{\textit{Ours} has lowest FIDs as it learns to depart from the model.}
	\corr{Conversely for \cite{jahanian2019steerability}, when}{Instead when} increasing $\lambda$\corr{ it}{, \cite{jahanian2019steerability}} learns to mimic model but FID diverges from real images features.}\label{fig:miniplots}

\end{figure}

\subsection{Ablation studies}
\label{sec:exp-ablation}
\paragraph{Architectural changes.}
We ablate the use of $\mathcal{L}_M$ and $\mathcal{L}_{\phi}$ by removing either. \corr{For evaluating translation diversity on $\text{Day}\mapsto\text{Timelapse}$}{To evaluate the diversity of $\text{Day}\mapsto\text{Timelapse}$ translations}, we sample 10 couples of random $\{\phi_1, \phi_2\}$ \corr{translations }{}for 100 images and evaluate the LPIPS distance among \corr{each pair of images}{translations pairs}. We obtain LPIPS 0.020 \textit{w/o} $\mathcal{L}_M$, 0.044 \textit{w/o} $\mathcal{L}_{\phi}$, while using both proves best with \textbf{0.236}.%

\paragraph{Disentangled reconstruction.}
While we disentangle real domain $Y(\phi)$ and model domain $Y_M({\phi})$ (cf. Fig.~\ref{fig:method}), steerable GANs~\cite{jahanian2019steerability} instead leverage guidance directly on $Y(\phi)$. To study either benefit, we replace $\mathcal{L}^{\phi}$ and $\mathcal{L}_M$ with $\mathcal{L}_{edit} = \lambda||y^\phi-\Tilde{y}^\phi_{M}||_1$ as in~\cite{jahanian2019steerability}. Fig.~\ref{fig:miniplots} shows discrete FIDs\corr{}{,} for \textit{ours} and~\cite{jahanian2019steerability} with $\lambda = {1, 5}$\corr{.  The FIDs are}{,} evaluated against real data (blue) or model translations (orange). 
The plots hold complex but interesting insights. Specifically, low FIDs at Dawn/Dusk infer the model is reliable there, while divergent FIDs at night mean the opposite. With $\lambda=1$ the i2i lacks guidance and performs poorly, but higher $\lambda$ increases model mimicking and lower \textit{real} FID. Instead, \textit{ours} is \textit{guided by the model} but learns to depart from it with the discovery of exclusive target features.

\paragraph{Model choice.}
We \corr{ablate}{study} the \corr{importance}{benefit} of FIN encoding by \corr{switching}{swapping} linear and cyclic. Comparing with Tab.~\ref{tab:gan-metrics}, training $\text{iPhone}\mapsto\text{DSLR}$ with \textit{cyclic FIN} \corr{behaves}{is} worse (IS/CIS/LPIPS 1.41/1.20/0.678) and at the cost of more complex encoding.
Training $\text{Day}\mapsto\text{Timelapse}$ with \textit{linear FIN} performs on par \corr{}{or better} (IS/CIS/LPIPS 1.65/1.64/0.579) but \textit{loses dusk/dawn distinction} capability.

\begin{figure}
    \centering
    \begin{subfigure}[t]{\linewidth}
   		\scriptsize
    	\setlength{\tabcolsep}{0.005\linewidth}
    	\begin{tabular}{cccc}
    		\includegraphics[width=0.24\linewidth]{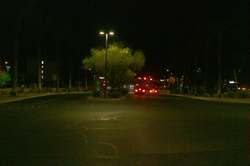}
    		& \includegraphics[width=0.24\linewidth]{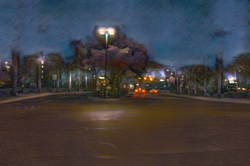}
    		& \includegraphics[width=0.24\linewidth]{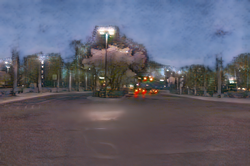}
    		& \includegraphics[width=0.24\linewidth]{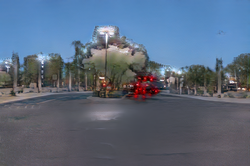}\\
    		Input ($\phi_{est}$)
    		& Rel. $+17.5\degree$
    		& Abs. to $-5\degree$ (dusk)
    		& Abs. to $30\degree$ (day)
    	\end{tabular}\vspace{-0.5em}
 		\caption{$\phi$-agnostic inference}
 		\label{fig:unsup-inference}
    \end{subfigure}
    \begin{subfigure}[t]{\linewidth}
		\resizebox{\linewidth}{!}{
			\setlength{\tabcolsep}{0.005\linewidth}
			\begin{tabular}{c | c c c c c c c c c}
				\includegraphics[width=7em, valign=m]{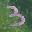}\hspace{0.5em}
				& \hspace{0.2em}\includegraphics[width=7em, valign=m]{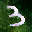}
				& \includegraphics[width=7em, valign=m]{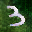}
				& \includegraphics[width=7em, valign=m]{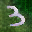}
				& \includegraphics[width=7em, valign=m]{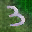}
				& \includegraphics[width=7em, valign=m]{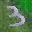}
				& \includegraphics[width=7em, valign=m]{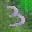}
				& \includegraphics[width=7em, valign=m]{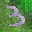}
				& \includegraphics[width=7em, valign=m]{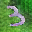}
				& \includegraphics[width=7em, valign=m]{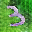}\\
				\includegraphics[width=7em, height=7em, valign=m]{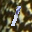}\hspace{0.2em}
				& \hspace{0.2em}\includegraphics[width=7em, height=7em, valign=m]{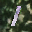}
				& \includegraphics[width=7em, height=7em, valign=m]{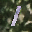}
				& \includegraphics[width=7em, height=7em, valign=m]{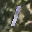}
				& \includegraphics[width=7em, height=7em, valign=m]{figures/mnistm/redness_im_10_3.png}
				& \includegraphics[width=7em, height=7em, valign=m]{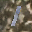}
				& \includegraphics[width=7em, height=7em, valign=m]{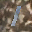}
				& \includegraphics[width=7em, height=7em, valign=m]{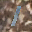}
				& \includegraphics[width=7em, height=7em, valign=m]{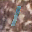}
				& \includegraphics[width=7em, height=7em, valign=m]{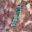}\\
				\huge{Source} & \tikzmark{a}{} &  &  &  &  &  &  &  & \tikzmark{b}{}\\
		\end{tabular}\link{a}{b}}\vspace{-0.5em}
		\caption{Training with domain confusion}
		\label{fig:qualitative-mnistm}
    \end{subfigure}
    \begin{subfigure}[t]{0.7\linewidth}
        \resizebox{\linewidth}{!}{
		\setlength{\tabcolsep}{0.005\linewidth}
       	\begin{tabular}{c | c c c}
            \includegraphics[width=7em]{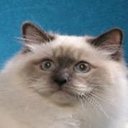}&
            \includegraphics[width=7em]{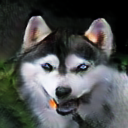}&
            \includegraphics[width=7em]{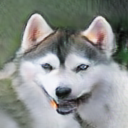}&
            \includegraphics[width=7em]{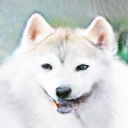}\vspace{-0.5em}\\
			Source &\tikzmark{a}{} &  & \tikzmark{b}{}\vspace{-0.2em}\\
			Cat & Dog (black fur) & \textbf{CoMo-MUNIT} & Dog (white fur)
        \end{tabular}\link{a}{b}}
        \caption{$\text{Cat}\mapsto\text{Dog}$ with fur color guidance}\label{fig:cat2dog}
    \end{subfigure}
	\caption{\protect\subref{fig:unsup-inference}: Training with shared encoder/decoder \corr{parameters }{}and using $\phi\text{-Net}_A$ at inference\corr{,}{} enables relative and absolute $\phi$ translations. The input is estimated at $\phi_{est} = -33.45\degree$ (gt $-32.73\degree$) and \corr{is }{}shifted with various strategies.
	\protect\subref{fig:qualitative-mnistm}: \corr{Sample translations of }{}$\text{CoMo-CycleGAN}$ on MNIST-M~\cite{ganin2016domain} trained with \textit{domain confusion}\corr{. Guidance is on brightness (1st row) or redness (2nd) and assumes no fixed $\phi$}{ (w/o fixed $\phi$), guiding on brightness (1st row) or redness (2nd)}. \corr{The figure}{It} shows source (leftmost) and translations along \corr{the }{}$\phi$ dimension. Despite domain confusion, it reorganized the manifold and produced valid translations. In \protect\subref{fig:cat2dog}, \corr{we guide fur color for a transformation where continuity is not immediate (\textit{dogness}?)}{we guide the complex $\text{Cat}\mapsto\text{Dog}$ only with fur color}.
    }
\end{figure}
\section{Discussion}
\label{sec:discussion}
\label{sec:exp-extension}
\corr{While we proved CoMoGAN is beneficial to a variety of continuous translations problems, we wish now to discuss possible extensions and limitations of the method.}{}

\paragraph{$\phi$-agnostic inference.}
In all experiments, translation assumes source at $\phi_0$, though agnostic inference is of interest. 
To test this, we trained our method with cycle consistency and shared parameters for $X\mapsto Y$ and $Y\mapsto X$ encoder/decoders (refer to Sec.~\ref{method:implementation}). At inference, we used $\phi\text{-Net}_A$ to estimate $\phi_{est}$ on input which enabled absolute translation regardless of input (e.g. anytime$\mapsto$day) but also relative translation (e.g.~$+5\degree$). Sample results in Fig.~\ref{fig:unsup-inference} show exciting results with challenging night input.

\paragraph{Source/Target domains confusion.} 
A limitation of most GANs is the need of source/target splits while \textit{truly unsupervised} GAN could discover a continuous manifold from mixed source/target data (i.e. $X \cup Y$ or domains confusion).
Interestingly, model-guided GANs \corr{like ours may }{}allow this \corr{if}{\textit{if}} the model \corr{is not limited to the $\phi$-nature of the input data}{does not enforce $\phi$ input}.
While there are no physical model for bilateral $\text{night}\leftrightarrow\text{day}$ or $\text{foggy}\leftrightarrow\text{clear}$, we prove the feasibility \corr{on toy tasks }on MNIST-M~\cite{ganin2016domain} \corr{}{ toy tasks, }learning \textit{brightness} or \textit{redness} manifold. \corr{Training only with one set, }{}Fig.~\ref{fig:qualitative-mnistm} shows we correctly achieve translation, \corr{opening doors towards unsupervised guidance in continuous domains}{paving ways for truly unsupervised GAN}.

\paragraph{Models and data limitations.}
\corr{It is worth noting that model-guided GAN may not be usable in some complex scenarios (e.g. face-to-face, etc.) due to the lack of sufficiently reliable models, still CoMoGAN can be used to guide features as skin tone, etc. as visible in the experiment in fig.~\ref{fig:cat2dog} on $\text{Cat}\mapsto\text{Dog}$ with fur color guidance.}{{Model-guided GAN are unsuitable for some complex scenarios (e.g. face-to-face) due to the lack of models, but can guide features as skin tone, etc. as in our experiment Fig.~\ref{fig:cat2dog} on $\text{Cat}\mapsto\text{Dog}$ using fur color guidance.}}
\corr{On the other hand, similar to the literature~\cite{jahanian2019steerability}, we experienced that data scarcity affects greatly the unsupervised manifold discovery. This was noted in experiments, removing dusk and dawn data to train timelapse translations, which proved to fail drastically.}{Like~\cite{jahanian2019steerability}, we too experienced that data scarcity affects greatly the manifold discovery and training timelapse without dusk and dawn proves to fail drastically.}

\paragraph{Acknowledgements}
This work used HPC resources from GENCI–IDRIS (Grant 2020-AD011012040).

{\small
\bibliographystyle{ieee_fullname}	
\bibliography{egbib}

\begin{thebibliography}{10}\itemsep=-1pt

\bibitem{pysolar}
Pysolar.
\newblock \url{https://github.com/pingswept/pysolar}.

\bibitem{almahairi2018augmented}
Amjad Almahairi, Sai Rajeswar, Alessandro Sordoni, Philip Bachman, and Aaron
  Courville.
\newblock Augmented cyclegan: Learning many-to-many mappings from unpaired
  data.
\newblock {\em ICML}, 2018.

\bibitem{anokhin2020high}
Ivan Anokhin, Pavel Solovev, Denis Korzhenkov, Alexey Kharlamov, Taras
  Khakhulin, Aleksei Silvestrov, Sergey Nikolenko, Victor Lempitsky, and Gleb
  Sterkin.
\newblock High-resolution daytime translation without domain labels.
\newblock In {\em CVPR}, 2020.

\bibitem{baek2020tunit}
Kyungjune Baek, Yunjey Choi, Youngjung Uh, Jaejun Yoo, and Hyunjung Shim.
\newblock Rethinking the truly unsupervised image-to-image translation.
\newblock {\em arXiv}, 2020.

\bibitem{bhattacharjee2020dunit}
Deblina Bhattacharjee, Seungryong Kim, Guillaume Vizier, and Mathieu Salzmann.
\newblock Dunit: Detection-based unsupervised image-to-image translation.
\newblock In {\em CVPR}, 2020.

\bibitem{chen2019homomorphic}
Ying-Cong Chen, Xiaogang Xu, Zhuotao Tian, and Jiaya Jia.
\newblock Homomorphic latent space interpolation for unpaired image-to-image
  translation.
\newblock In {\em CVPR}, 2019.

\bibitem{cherian2019sem}
Anoop Cherian and Alan Sullivan.
\newblock Sem-gan: Semantically-consistent image-to-image translation.
\newblock In {\em WACV}, 2019.

\bibitem{choi2018stargan}
Yunjey Choi, Minje Choi, Munyoung Kim, Jung-Woo Ha, Sunghun Kim, and Jaegul
  Choo.
\newblock Stargan: Unified generative adversarial networks for multi-domain
  image-to-image translation.
\newblock In {\em CVPR}, 2018.

\bibitem{choi2020stargan}
Yunjey Choi, Youngjung Uh, Jaejun Yoo, and Jung-Woo Ha.
\newblock Stargan v2: Diverse image synthesis for multiple domains.
\newblock In {\em CVPR}, 2020.

\bibitem{cordts2016cityscapes}
Marius Cordts, Mohamed Omran, Sebastian Ramos, Timo Rehfeld, Markus Enzweiler,
  Rodrigo Benenson, Uwe Franke, Stefan Roth, and Bernt Schiele.
\newblock The cityscapes dataset for semantic urban scene understanding.
\newblock In {\em CVPR}, 2016.

\bibitem{ganin2016domain}
Yaroslav Ganin, Evgeniya Ustinova, Hana Ajakan, Pascal Germain, Hugo
  Larochelle, Fran{\c{c}}ois Laviolette, Mario Marchand, and Victor Lempitsky.
\newblock Domain-adversarial training of neural networks.
\newblock {\em JMLR}, 2016.

\bibitem{goetschalckx2019ganalyze}
Lore Goetschalckx, Alex Andonian, Aude Oliva, and Phillip Isola.
\newblock Ganalyze: Toward visual definitions of cognitive image properties.
\newblock In {\em ICCV}, 2019.

\bibitem{gong2020analogical}
Rui Gong, Dengxin Dai, Yuhua Chen, Wen Li, and Luc Van~Gool.
\newblock Analogical image translation for fog generation.
\newblock In {\em AAAI}, 2021.

\bibitem{gong2019dlow}
Rui Gong, Wen Li, Yuhua Chen, and Luc~Van Gool.
\newblock Dlow: Domain flow for adaptation and generalization.
\newblock In {\em CVPR}, 2019.

\bibitem{halder2019physics}
Shirsendu~Sukanta Halder, Jean-Fran{\c{c}}ois Lalonde, and Raoul de Charette.
\newblock Physics-based rendering for improving robustness to rain.
\newblock In {\em ICCV}, 2019.

\bibitem{he2016deep}
Kaiming He, Xiangyu Zhang, Shaoqing Ren, and Jian Sun.
\newblock Deep residual learning for image recognition.
\newblock In {\em CVPR}, 2016.

\bibitem{heusel2017gans}
Martin Heusel, Hubert Ramsauer, Thomas Unterthiner, Bernhard Nessler, and Sepp
  Hochreiter.
\newblock Gans trained by a two time-scale update rule converge to a local nash
  equilibrium.
\newblock In {\em NeurIPS}, 2017.

\bibitem{hoffman2017cycada}
Judy Hoffman, Eric Tzeng, Taesung Park, and Phillip~Isola Jun-Yan~Zhu, Kate
  Saenko, Alexei~A. Efros, and Trevor Darrell.
\newblock Cycada: Cycle consistent adversarial domain adaptation.
\newblock In {\em ICML}, 2018.

\bibitem{hosek2012analytic}
Lukas Hosek and Alexander Wilkie.
\newblock An analytic model for full spectral sky-dome radiance.
\newblock {\em TOG}, 2012.

\bibitem{huang2017arbitrary}
Xun Huang and Serge Belongie.
\newblock Arbitrary style transfer in real-time with adaptive instance
  normalization.
\newblock In {\em ICCV}, 2017.

\bibitem{huang2018multimodal}
Xun Huang, Ming-Yu Liu, Serge Belongie, and Jan Kautz.
\newblock Multimodal unsupervised image-to-image translation.
\newblock In {\em ECCV}, 2018.

\bibitem{isola2017image}
Phillip Isola, Jun-Yan Zhu, Tinghui Zhou, and Alexei~A Efros.
\newblock Image-to-image translation with conditional adversarial networks.
\newblock In {\em CVPR}, 2017.

\bibitem{jahanian2019steerability}
Ali Jahanian, Lucy Chai, and Phillip Isola.
\newblock On the "steerability" of generative adversarial networks.
\newblock In {\em ICLR}, 2020.

\bibitem{jiang2020tsit}
Liming Jiang, Changxu Zhang, Mingyang Huang, Chunxiao Liu, Jianping Shi, and
  Chen~Change Loy.
\newblock Tsit: A simple and versatile framework for image-to-image
  translation.
\newblock In {\em ECCV}, 2020.

\bibitem{Karpatne2017PhysicsguidedNN}
A. Karpatne, William Watkins, J. Read, and V. Kumar.
\newblock Physics-guided neural networks (pgnn): An application in lake
  temperature modeling.
\newblock {\em arXiv}, 2017.

\bibitem{lee2019drit++}
Hsin-Ying Lee, Hung-Yu Tseng, Qi Mao, Jia-Bin Huang, Yu-Ding Lu, Maneesh Singh,
  and Ming-Hsuan Yang.
\newblock Drit++: Diverse image-to-image translation via disentangled
  representations.
\newblock {\em IJCV}, 2020.

\bibitem{li2018semantic}
Peilun Li, Xiaodan Liang, Daoyuan Jia, and Eric~P Xing.
\newblock Semantic-aware grad-gan for virtual-to-real urban scene adaption.
\newblock {\em BMVC}, 2018.

\bibitem{li2019heavy}
Ruoteng Li, Loong-Fah Cheong, and Robby~T Tan.
\newblock Heavy rain image restoration: Integrating physics model and
  conditional adversarial learning.
\newblock In {\em CVPR}, 2019.

\bibitem{li2019bidirectional}
Yunsheng Li, Lu Yuan, and Nuno Vasconcelos.
\newblock Bidirectional learning for domain adaptation of semantic
  segmentation.
\newblock In {\em CVPR}, 2019.

\bibitem{lin2020multimodal}
Che-Tsung Lin, Yen-Yi Wu, Po-Hao Hsu, and Shang-Hong Lai.
\newblock Multimodal structure-consistent image-to-image translation.
\newblock In {\em AAAI}, 2020.

\bibitem{lin2019exploring}
Jianxin Lin, Zhibo Chen, Yingce Xia, Sen Liu, Tao Qin, and Jiebo Luo.
\newblock Exploring explicit domain supervision for latent space
  disentanglement in unpaired image-to-image translation.
\newblock {\em T-PAMI}, 2019.

\bibitem{liu2017unsupervised}
Ming-Yu Liu, Thomas Breuel, and Jan Kautz.
\newblock Unsupervised image-to-image translation networks.
\newblock In {\em NeurIPS}, 2017.

\bibitem{liu2019few}
Ming-Yu Liu, Xun Huang, Arun Mallya, Tero Karras, Timo Aila, Jaakko Lehtinen,
  and Jan Kautz.
\newblock Few-shot unsupervised image-to-image translation.
\newblock In {\em ICCV}, 2019.

\bibitem{lu2019unsupervised}
Boyu Lu, Jun-Cheng Chen, and Rama Chellappa.
\newblock Unsupervised domain-specific deblurring via disentangled
  representations.
\newblock In {\em CVPR}, 2019.

\bibitem{lutjens2020physics}
Bj{\"o}rn L{\"u}tjens, Brandon Leshchinskiy, Christian Requena-Mesa, Farrukh
  Chishtie, Natalia D{\'\i}az-Rodriguez, Oc{\'e}ane Boulais, Aaron Pi{\~n}a,
  Dava Newman, Alexander Lavin, Yarin Gal, et~al.
\newblock Physics-informed gans for coastal flood visualization.
\newblock {\em arXiv}, 2020.

\bibitem{ma2018exemplar}
Liqian Ma, Xu Jia, Stamatios Georgoulis, Tinne Tuytelaars, and Luc Van~Gool.
\newblock Exemplar guided unsupervised image-to-image translation with semantic
  consistency.
\newblock In {\em ICLR}, 2019.

\bibitem{mao2020continuous}
Qi Mao, Hsin-Ying Lee, Hung-Yu Tseng, Jia-Bin Huang, Siwei Ma, and Ming-Hsuan
  Yang.
\newblock Continuous and diverse image-to-image translation via signed
  attribute vectors.
\newblock {\em arXiv}, 2020.

\bibitem{mao2017least}
Xudong Mao, Qing Li, Haoran Xie, Raymond~YK Lau, Zhen Wang, and Stephen
  Paul~Smolley.
\newblock Least squares generative adversarial networks.
\newblock In {\em ICCV}, 2017.

\bibitem{mo2018instagan}
Sangwoo Mo, Minsu Cho, and Jinwoo Shin.
\newblock Instagan: Instance-aware image-to-image translation.
\newblock {\em ICLR}, 2019.

\bibitem{murez2018image}
Zak Murez, Soheil Kolouri, David Kriegman, Ravi Ramamoorthi, and Kyungnam Kim.
\newblock Image to image translation for domain adaptation.
\newblock In {\em CVPR}, 2018.

\bibitem{musto2020semantically}
Luigi Musto and Andrea Zinelli.
\newblock Semantically adaptive image-to-image translation for domain
  adaptation of semantic segmentation.
\newblock In {\em BMVC}, 2020.

\bibitem{nguyen2019hologan}
Thu Nguyen-Phuoc, Chuan Li, Lucas Theis, Christian Richardt, and Yong-Liang
  Yang.
\newblock Hologan: Unsupervised learning of 3d representations from natural
  images.
\newblock In {\em ICCV}, 2019.

\bibitem{pan2018physics}
Jinshan Pan, Jiangxin Dong, Yang Liu, Jiawei Zhang, Jimmy Ren, Jinhui Tang,
  Yu-Wing Tai, and Ming-Hsuan Yang.
\newblock Physics-based generative adversarial models for image restoration and
  beyond.
\newblock {\em T-PAMI}, 2018.

\bibitem{park2020contrastive}
Taesung Park, Alexei~A Efros, Richard Zhang, and Jun-Yan Zhu.
\newblock Contrastive learning for unpaired image-to-image translation.
\newblock In {\em ECCV}, 2020.

\bibitem{park2019semantic}
Taesung Park, Ming-Yu Liu, Ting-Chun Wang, and Jun-Yan Zhu.
\newblock Semantic image synthesis with spatially-adaptive normalization.
\newblock In {\em CVPR}, 2019.

\bibitem{pizzati2020model}
Fabio Pizzati, Pietro Cerri, and Raoul de Charette.
\newblock Model-based occlusion disentanglement for image-to-image translation.
\newblock In {\em ECCV}, 2020.

\bibitem{pizzati2019domain}
Fabio Pizzati, Raoul de Charette, Michela Zaccaria, and Pietro Cerri.
\newblock Domain bridge for unpaired image-to-image translation and
  unsupervised domain adaptation.
\newblock In {\em WACV}, 2020.

\bibitem{pumarola2020ganimation}
Albert Pumarola, Antonio Agudo, Aleix~M Martinez, Alberto Sanfeliu, and
  Francesc Moreno-Noguer.
\newblock Ganimation: One-shot anatomically consistent facial animation.
\newblock {\em IJCV}, 2020.

\bibitem{reichstein2019deep}
Markus Reichstein, Gustau Camps-Valls, Bjorn Stevens, Martin Jung, Joachim
  Denzler, Nuno Carvalhais, et~al.
\newblock Deep learning and process understanding for data-driven earth system
  science.
\newblock {\em Nature}, 2019.

\bibitem{romero2019smit}
Andr{\'e}s Romero, Pablo Arbel{\'a}ez, Luc Van~Gool, and Radu Timofte.
\newblock Smit: Stochastic multi-label image-to-image translation.
\newblock In {\em ICCV Workshops}, 2019.

\bibitem{ros2016synthia}
German Ros, Laura Sellart, Joanna Materzynska, David Vazquez, and Antonio~M
  Lopez.
\newblock The synthia dataset: A large collection of synthetic images for
  semantic segmentation of urban scenes.
\newblock In {\em CVPR}, 2016.

\bibitem{saito2020coco}
Kuniaki Saito, Kate Saenko, and Ming-Yu Liu.
\newblock Coco-funit: Few-shot unsupervised image translation with a content
  conditioned style encoder.
\newblock In {\em ECCV}, 2020.

\bibitem{sakaridis2018semantic}
Christos Sakaridis, Dengxin Dai, and Luc Van~Gool.
\newblock Semantic foggy scene understanding with synthetic data.
\newblock {\em IJCV}, 2018.

\bibitem{salimans2016improved}
Tim Salimans, Ian Goodfellow, Wojciech Zaremba, Vicki Cheung, Alec Radford, and
  Xi Chen.
\newblock Improved techniques for training gans.
\newblock In {\em NeurIPS}, 2016.

\bibitem{shen2019towards}
Zhiqiang Shen, Mingyang Huang, Jianping Shi, Xiangyang Xue, and Thomas~S Huang.
\newblock Towards instance-level image-to-image translation.
\newblock In {\em CVPR}, 2019.

\bibitem{singh2019finegan}
Krishna~Kumar Singh, Utkarsh Ojha, and Yong~Jae Lee.
\newblock Finegan: Unsupervised hierarchical disentanglement for fine-grained
  object generation and discovery.
\newblock In {\em CVPR}, 2019.

\bibitem{sun2020scalability}
Pei Sun, Henrik Kretzschmar, Xerxes Dotiwalla, Aurelien Chouard, Vijaysai
  Patnaik, Paul Tsui, James Guo, Yin Zhou, Yuning Chai, Benjamin Caine, et~al.
\newblock Scalability in perception for autonomous driving: Waymo open dataset.
\newblock In {\em CVPR}, 2020.

\bibitem{tang2020multi}
Hao Tang, Dan Xu, Yan Yan, Jason~J Corso, Philip~HS Torr, and Nicu Sebe.
\newblock Multi-channel attention selection gans for guided image-to-image
  translation.
\newblock In {\em CVPR}, 2019.

\bibitem{thompson2002spatial}
William~B Thompson, Peter Shirley, and James~A Ferwerda.
\newblock A spatial post-processing algorithm for images of night scenes.
\newblock {\em Journal of Graphics Tools}, 2002.

\bibitem{toldo2020unsupervised}
Marco Toldo, Umberto Michieli, Gianluca Agresti, and Pietro Zanuttigh.
\newblock Unsupervised domain adaptation for mobile semantic segmentation based
  on cycle consistency and feature alignment.
\newblock {\em Image and Vision Computing}, 2020.

\bibitem{tremblay2020rain}
Maxime Tremblay, Shirsendu~Sukanta Halder, Raoul de Charette, and
  Jean-Fran{\c{c}}ois Lalonde.
\newblock Rain rendering for evaluating and improving robustness to bad
  weather.
\newblock {\em IJCV}, 2020.

\bibitem{ulyanov2017improved}
Dmitry Ulyanov, Andrea Vedaldi, and Victor Lempitsky.
\newblock Improved texture networks: Maximizing quality and diversity in
  feed-forward stylization and texture synthesis.
\newblock In {\em CVPR}, 2017.

\bibitem{upchurch2017deep}
Paul Upchurch, Jacob Gardner, Geoff Pleiss, Robert Pless, Noah Snavely, Kavita
  Bala, and Kilian Weinberger.
\newblock Deep feature interpolation for image content changes.
\newblock In {\em CVPR}, 2017.

\bibitem{wang2019deep}
Xintao Wang, Ke Yu, Chao Dong, Xiaoou Tang, and Chen~Change Loy.
\newblock Deep network interpolation for continuous imagery effect transition.
\newblock In {\em CVPR}, 2019.

\bibitem{wu2019relgan}
Po-Wei Wu, Yu-Jing Lin, Che-Han Chang, Edward~Y Chang, and Shih-Wei Liao.
\newblock Relgan: Multi-domain image-to-image translation via relative
  attributes.
\newblock In {\em ICCV}, 2019.

\bibitem{xia2020unsupervised}
Weihao Xia, Yujiu Yang, and Jing-Hao Xue.
\newblock Unsupervised multi-domain multimodal image-to-image translation with
  explicit domain-constrained disentanglement.
\newblock {\em Neural Networks}, 2020.

\bibitem{xiao2017dna}
Taihong Xiao, Jiapeng Hong, and Jinwen Ma.
\newblock Dna-gan: Learning disentangled representations from multi-attribute
  images.
\newblock {\em ICLR Workshops}, 2018.

\bibitem{xie2018tempogan}
You Xie, Erik Franz, Mengyu Chu, and Nils Thuerey.
\newblock tempogan: A temporally coherent, volumetric gan for super-resolution
  fluid flow.
\newblock {\em SIGGRAPH}, 2018.

\bibitem{yang2018towards}
Xitong Yang, Zheng Xu, and Jiebo Luo.
\newblock Towards perceptual image dehazing by physics-based disentanglement
  and adversarial training.
\newblock In {\em AAAI}, 2018.

\bibitem{yi2017dualgan}
Zili Yi, Hao Zhang, Ping Tan, and Minglun Gong.
\newblock Dualgan: Unsupervised dual learning for image-to-image translation.
\newblock In {\em ICCV}, 2017.

\bibitem{zhang2019multi}
Jianfu Zhang, Yuanyuan Huang, Yaoyi Li, Weijie Zhao, and Liqing Zhang.
\newblock Multi-attribute transfer via disentangled representation.
\newblock In {\em AAAI}, 2019.

\bibitem{zhang2018unreasonable}
Richard Zhang, Phillip Isola, Alexei~A Efros, Eli Shechtman, and Oliver Wang.
\newblock The unreasonable effectiveness of deep features as a perceptual
  metric.
\newblock In {\em CVPR}, 2018.

\bibitem{zhao2017pyramid}
Hengshuang Zhao, Jianping Shi, Xiaojuan Qi, Xiaogang Wang, and Jiaya Jia.
\newblock Pyramid scene parsing network.
\newblock In {\em CVPR}, 2017.

\bibitem{zheng2020forkgan}
Ziqiang Zheng, Yang Wu, Xinran Han, and Jianbo Shi.
\newblock Forkgan: Seeing into the rainy night.
\newblock In {\em ECCV}, 2020.

\bibitem{zhu2017unpaired}
Jun-Yan Zhu, Taesung Park, Phillip Isola, and Alexei~A Efros.
\newblock Unpaired image-to-image translation using cycle-consistent
  adversarial networks.
\newblock In {\em CVPR}, 2017.

\bibitem{zhu2017toward}
Jun-Yan Zhu, Richard Zhang, Deepak Pathak, Trevor Darrell, Alexei~A Efros,
  Oliver Wang, and Eli Shechtman.
\newblock Toward multimodal image-to-image translation.
\newblock In {\em NeurIPS}, 2017.

\bibitem{zhu2020sean}
Peihao Zhu, Rameen Abdal, Yipeng Qin, and Peter Wonka.
\newblock Sean: Image synthesis with semantic region-adaptive normalization.
\newblock In {\em CVPR}, 2020.

\bibitem{zhu2020semantically}
Zhen Zhu, Zhiliang Xu, Ansheng You, and Xiang Bai.
\newblock Semantically multi-modal image synthesis.
\newblock In {\em CVPR}, 2020.

\end{thebibliography}


\begin{thebibliography}{10}\itemsep=-1pt

\bibitem{choi2020stargan}
Yunjey Choi, Youngjung Uh, Jaejun Yoo, and Jung-Woo Ha.
\newblock Stargan v2: Diverse image synthesis for multiple domains.
\newblock In {\em CVPR}, 2020.

\bibitem{cordts2016cityscapes}
Marius Cordts, Mohamed Omran, Sebastian Ramos, Timo Rehfeld, Markus Enzweiler,
  Rodrigo Benenson, Uwe Franke, Stefan Roth, and Bernt Schiele.
\newblock The cityscapes dataset for semantic urban scene understanding.
\newblock In {\em CVPR}, 2016.

\bibitem{gong2019dlow}
Rui Gong, Wen Li, Yuhua Chen, and Luc~Van Gool.
\newblock Dlow: Domain flow for adaptation and generalization.
\newblock In {\em CVPR}, 2019.

\bibitem{halder2019physics}
Shirsendu~Sukanta Halder, Jean-Fran{\c{c}}ois Lalonde, and Raoul de Charette.
\newblock Physics-based rendering for improving robustness to rain.
\newblock In {\em ICCV}, 2019.

\bibitem{hosek2012analytic}
Lukas Hosek and Alexander Wilkie.
\newblock An analytic model for full spectral sky-dome radiance.
\newblock {\em TOG}, 2012.

\bibitem{huang2018multimodal}
Xun Huang, Ming-Yu Liu, Serge Belongie, and Jan Kautz.
\newblock Multimodal unsupervised image-to-image translation.
\newblock In {\em ECCV}, 2018.

\bibitem{jahanian2019steerability}
Ali Jahanian, Lucy Chai, and Phillip Isola.
\newblock On the "steerability" of generative adversarial networks.
\newblock In {\em ICLR}, 2020.

\bibitem{ros2016synthia}
German Ros, Laura Sellart, Joanna Materzynska, David Vazquez, and Antonio~M
  Lopez.
\newblock The synthia dataset: A large collection of synthetic images for
  semantic segmentation of urban scenes.
\newblock In {\em CVPR}, 2016.

\bibitem{thompson2002spatial}
William~B Thompson, Peter Shirley, and James~A Ferwerda.
\newblock A spatial post-processing algorithm for images of night scenes.
\newblock {\em Journal of Graphics Tools}, 2002.

\bibitem{tremblay2020rain}
Maxime Tremblay, Shirsendu~Sukanta Halder, Raoul de Charette, and
  Jean-Fran{\c{c}}ois Lalonde.
\newblock Rain rendering for evaluating and improving robustness to bad
  weather.
\newblock {\em IJCV}, 2020.

\bibitem{wang2019deep}
Xintao Wang, Ke Yu, Chao Dong, Xiaoou Tang, and Chen~Change Loy.
\newblock Deep network interpolation for continuous imagery effect transition.
\newblock In {\em CVPR}, 2019.

\bibitem{zhu2017unpaired}
Jun-Yan Zhu, Taesung Park, Phillip Isola, and Alexei~A Efros.
\newblock Unpaired image-to-image translation using cycle-consistent
  adversarial networks.
\newblock In {\em CVPR}, 2017.

\end{thebibliography}
}

\end{document}